\documentclass[conference]{IEEEtran}
\IEEEoverridecommandlockouts
\usepackage{cite}
\usepackage{amsmath,amssymb,amsfonts}
\usepackage{algorithmic}
\usepackage{graphicx}
\usepackage{multirow}
\usepackage{textcomp}
\usepackage{xcolor}
\usepackage{todonotes}
\usepackage{hyperref}
\usepackage{booktabs}
\usepackage{wrapfig}
\usepackage{subcaption}

\def\BibTeX{{\rm B\kern-.05em{\sc i\kern-.025em b}\kern-.08em
    T\kern-.1667em\lower.7ex\hbox{E}\kern-.125emX}}
\begin{document}

\title{Gym-$\mu$RTS: Toward Affordable Full Game Real-time Strategy Games Research with Deep Reinforcement Learning
}

\author{\IEEEauthorblockN{Shengyi Huang}
\IEEEauthorblockA{
\textit{Drexel University}\\
Philadelphia, PA, USA \\
sh3397@drexel.edu}
\and
\IEEEauthorblockN{Santiago Onta\~{n}\'{o}n* \thanks{*Currently at Google}}
\IEEEauthorblockA{
\textit{Drexel University}\\
Philadelphia, PA, USA \\
sh3397@drexel.edu}
\and
\IEEEauthorblockN{Chris Bamford}
\IEEEauthorblockA{ 
\textit{Queen Mary University}\\
London, United Kindom \\
c.d.j.bamford@qmul.ac.uk}
\and
\IEEEauthorblockN{Lukasz Grela}
\IEEEauthorblockA{ 
grotrianster@gmail.com}
}

\IEEEoverridecommandlockouts
\IEEEpubid{\makebox[\columnwidth]{978-1-6654-3886-5/21/\$31.00 ©2021 IEEE \hfill} \hspace{\columnsep}\makebox[\columnwidth]{ }}

\maketitle

\begin{abstract}

In recent years, researchers have achieved great success in applying Deep Reinforcement Learning (DRL) algorithms to Real-time Strategy (RTS) games, creating strong autonomous agents that could defeat professional players in StarCraft~II. However, existing approaches to tackle full games have high computational costs, usually requiring the use of thousands of GPUs and CPUs for weeks. This paper has two main contributions to address this issue: 1) We introduce Gym-$\mu$RTS (pronounced ``gym-micro-RTS'') as a fast-to-run RL environment for full-game RTS research and 2) we present a collection of techniques to scale DRL to play full-game $\mu$RTS as well as ablation studies to demonstrate their empirical importance. Our best-trained bot can defeat every $\mu$RTS bot we tested from the past $\mu$RTS competitions when working in a single-map setting, resulting in a state-of-the-art DRL agent while only taking about 60 hours of training using a single machine (one GPU, three vCPU, 16GB RAM).



\end{abstract}


\section{Introduction}
In recent years, researchers have achieved great success in applying Deep Reinforcement Learning (DRL) algorithms to Real-time Strategy (RTS) games. Most notably, DeepMind trained a grandmaster-level AI called AlphaStar with DRL for the popular RTS game StarCraft II~\cite{vinyals_2019}. AlphaStar demonstrates impressive strategy and game control, presenting many human-like behaviors, and is able to defeat professional players consistently. Given most previously designed bots fail to perform well in the full-game against humans~\cite{ontanon2013survey}, AlphaStar clearly represents a significant milestone in the field. 
%
While this accomplishment is impressive, it comes with high computational costs. In particular, AlphaStar and even further attempts by other teams to lower the computational costs~\cite{han2020tstarbot}  still require thousands of CPUs and GPUs/TPUs to train the agents for an extended period of time, which is outside of the computational budget of most researchers.

This paper has two main contributions to address this issue.
The first main contribution is to introduce Gym-$\mu$RTS  as an RL testbed for affordable full-game RTS research, which focuses on all aspects of the game such as harvesting resources, defending units, and attack enemies (this is in contrast to \emph{mini-games} that only focus on one aspect of the game). Gym-$\mu$RTS is a reinforcement learning interface for the RTS game $\mu$RTS~\cite{ontanon2013combinatorial}, which has been a popular platform to test out a variety of AI techniques for RTS games. Despite its simple visuals, $\mu$RTS captures the core challenges of RTS games. 
Although Gym-$\mu$RTS shares many similarities to the StarCraft II Learning Environment (PySC2)~\cite{vinyals2017starcraft}, there are also many key differences (e.g., PySC2 uses a human-like action space whereas Gym-$\mu$RTS uses a lower-level action space).
Through Gym-$\mu$RTS, we are able to conduct full-game RTS research using DRL without extensive technical resources such as high-performance compute clusters.

Despite the simplifications done in $\mu$RTS, playing 1v1 competitive matches via DRL is still a daunting task. Thus, our second main contribution is a collection of techniques to scale DRL to play $\mu$RTS. We start with a Proximal Policy Optimization (PPO)~\cite{schulman_2017} implementation that matches implementation details of PPO in \emph{openai/baselines}~\cite{baselines}, and incrementally stack augmentations to account for Gym-$\mu$RTS's combinatorial action space (all units must be controlled simultaneously) and improve training efficiency and performance. Among these augmentations, two are \emph{essential}: 1) action composition and 2) invalid action masking. These two augmentations combined allowed us to bootstrap an initial agent that could compete on the $16\times16$ map to a reasonable standard. Additionally, we experimented with 3) diversified training opponents, and 4) different neural network architectures. We provide ablation studies to shed insights on the importance of each of these augmentations. Our best-trained agent can defeat every $\mu$RTS bot we tested against, from the past $\mu$RTS competitions\footnote{\url{https://sites.google.com/site/micrortsaicompetition/home}} in a single-map setting, establishing a new state-of-the-art for DRL bots in $\mu$RTS while only taking about 60 hours of training using a single machine (one GPU, three vCPU, 16GB RAM).
We make source code and trained models \footnote{
\url{https://github.com/vwxyzjn/gym-microrts-paper} 
}, as well as all the metrics, logs, and recorded videos\footnote{
\url{https://wandb.ai/vwxyzjn/gym-microrts-paper}
} available for comparison. 

\begin{figure}[t]
\centering
    \includegraphics[width=0.7\columnwidth]{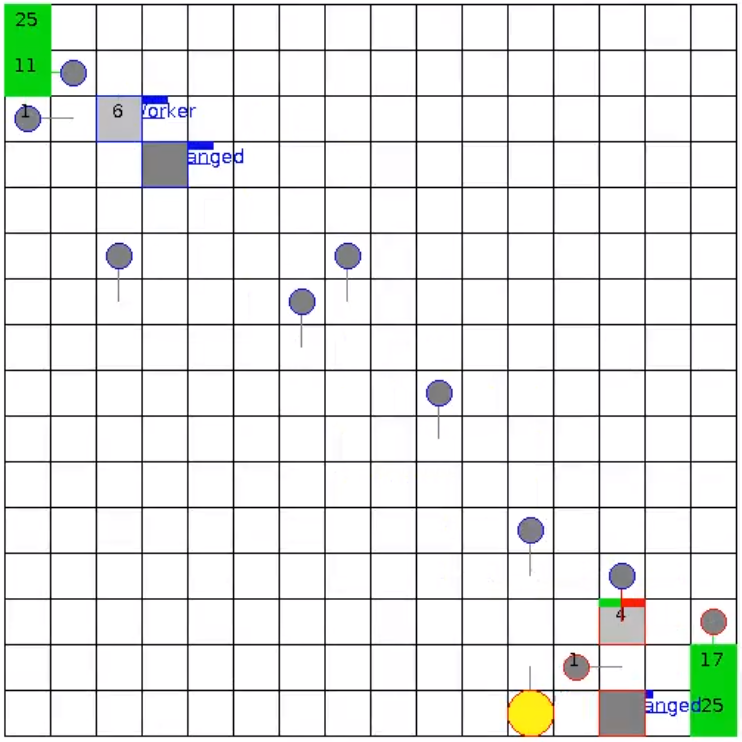}
  \caption{Screenshot of our best-trained agent (top-left) playing against CoacAI (bottom-right), the 2020 $\mu$RTS AI competition champion.  Strategy-wise, our agent usually defeats CoacAI by harvesting resources (green squares) efficiently using two workers (dark gray circles), doing a highly optimized worker rush that takes out the enemy base in the bottom right (shown with 50\% damage), followed by a transition to the mid and late game by producing combat units (colored circles) from the barracks (dark gray squares). 
  The blue and red border suggest the unit is owned by player 1 and 2, respectively. See additional combat videos here: \url{https://bit.ly/3llOhex}}
  \label{fig:WorkerRushAtEightBase}
\end{figure}

\section{Background}
Real-time Strategy (RTS) games are complex adversarial domains, typically simulating battles between a large number of combat units, that pose a significant challenge to both human and artificial intelligence \cite{buro2003real}. Designing AI techniques for RTS games is challenging  due to a variety of reasons: 
1) players need to issue actions in real-time, leaving little time computational budget, 2) the action spaces grows combinatorially with the number of units in the game, 3) the rewards are very sparse (win/loss at the end of the game), 4) generalizing against diverse set of opponents and maps is difficult, and 5) stochasticity of game mechanics and partial observability (these last two are not considered in this paper).


StarCraft I \& II are very popular RTS games and, among other games, have attracted much research attention. Past work in this area includes reinforcement learning~\cite{jaidee2012classq}, case-based reasoning~\cite{weber2009case,ontanon2010line}, or game tree search~\cite{balla2009uct,churchill2012fast,justesen2014script,ontanon2017combinatorial} among many other techniques designed to tackle different sub-problems in the game, such as micromanagement, or build-order generation. In the full-game settings, however, most techniques have had limited success in creating viable agents to play competitively against professional StarCraft players until recently. In particular, DeepMind introduced AlphaStar~\cite{vinyals_2019}, an agent trained with DRL and self-play, that sets a new state-of-the-art bot for StarCraft II, defeating professional players in the full-game. In Dota 2, a popular collaborative online-player game that shares many similar challenges as StarCraft, Open AI Five~\cite{Berner2019Dota2W} is able to create agents that can achieve super-human performance.
Although these two systems achieve great performance, they come with large computational costs. AlphaStar used 3072 TPU cores and 50,400 preemptible CPU cores for a duration of 44 days~\cite{vinyals_2019,han2020tstarbot}. This makes it difficult for those with less computational resources to do full-game RTS research using DRL.

There are usually three ways to circumvent this computational costs. The first way is to focus on sub problems such as combat scenarios~\cite{samvelyan2019starcraft}. The second way is to reduce the full-game complexity by either considering hierarchical actions spaces or incorporating scripted actions~\cite{sun2018tstarbots,lee2018modular}. The third way is to use alternative game simulators that run faster such as Mini-RTS~\cite{tian2017elf}, Deep RTS~\cite{andersen2018deep}, and CodeCraft\footnote{\url{https://github.com/cswinter/DeepCodeCraft}}. 

We show that Gym-$\mu$RTS as an alternative that could be used for full-game RTS research with the full action space while using affordable computational resources.

\section{Preliminaries}
In this paper, we use  policy gradient methods to train agents. 
Let us consider the Reinforcement Learning problem in a Markov Decision Process (MDP) denoted as $(S,A,P, \rho_0, r,\gamma, T)$, where $S$ is the state space, $A$ is the discrete action space, $P: S \times A \times S \rightarrow [0, 1]$ is the state transition probability, $\rho_0: S\rightarrow [0,1]$ is the initial state distribution, $r: S \times A \rightarrow \mathbb{R}$ is the reward function, $\gamma$ is the discount factor, and $T$ is the maximum episode length. A stochastic policy $\pi_{\theta}: S \times A \rightarrow [0,1]$, parameterized by a parameter vector $\theta$, assigns a probability value to an action given a state. The goal is to maximize the expected discounted return of the policy:
\begin{gather*}
        J = \mathbb{E}_{\tau}\left[\sum_{t=0}^{T-1} \gamma^{t} r_{t}\right]\\
        \text { where } \tau \text { is the trajectory } \left(s_{0}, a_{0}, r_{0},  \dots, s_{T-1}, a_{T-1}, r_{T-1}\right)\\
        \text { and } s_{0} \sim \rho_{0}, s_t \sim P(\cdot \vert s_{t-1}, a_{t-1}), a_t \sim \pi_{\theta}(\cdot \vert s_t), r_{t}=r\left(s_{t}, a_{t}\right)
\end{gather*}
The core idea behind policy gradient algorithms is to obtain the \textsl{policy gradient} $\nabla_{\theta}J$  of the expected discounted return with respect to the policy parameter $\theta$. Doing gradient ascent $\theta = \theta + \nabla_{\theta}J$ therefore maximizes the expected discounted reward. 
Earlier work proposes the following policy gradient estimate to the objective $J$~\cite{sutton2018reinforcement}:
\begin{align*}
   \nabla_{\theta}J = \mathbb{E}_{\tau\sim\pi_\theta}\left[ \sum_{t=0}^{T-1} \nabla_{\theta}\log\pi_{\theta}(a_t|s_t)G_t \right]\mathrm{,} \  G_{t} = \sum_{k=0}^{\infty} \gamma^{k} r_{t+k}
\end{align*}
This gradient estimate, however, suffers from high variance~\cite{sutton2018reinforcement} and there are many techniques and extensions to address it\cite{schulman2015high,schulman_2017,espeholt2018impala}.

\begin{figure}[t]
    \centering
    \includegraphics[width=0.8\columnwidth]{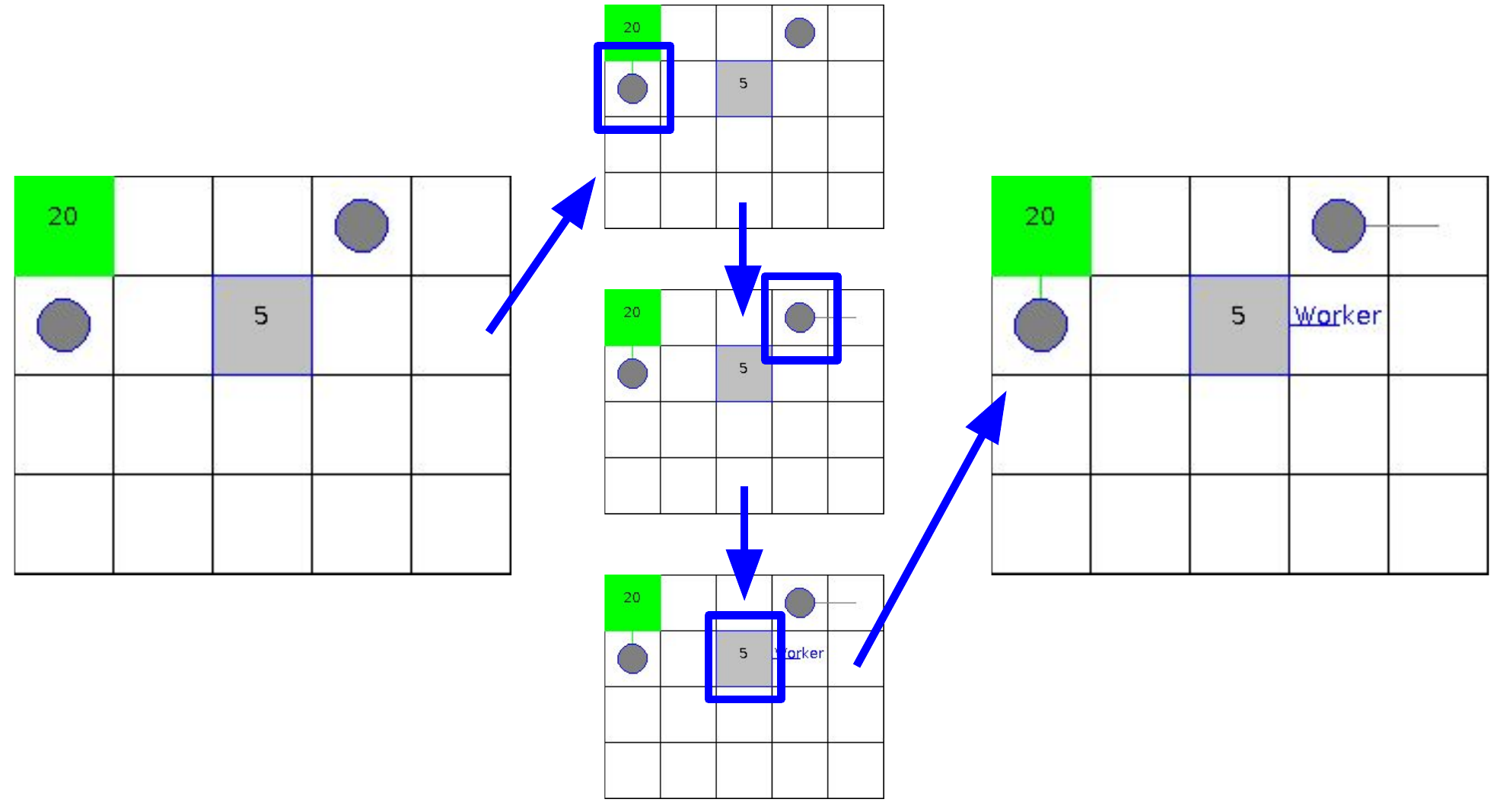}
    \caption{Unit Action Simulation (UAS) calls the policy iteratively until all units have actions (at each step, the policy chooses a unit and issue a unit action to it).}
    \label{fig:unitaActionSimulation}
\end{figure}
\begin{figure}[t]
    \centering
    \includegraphics[width=0.8\columnwidth]{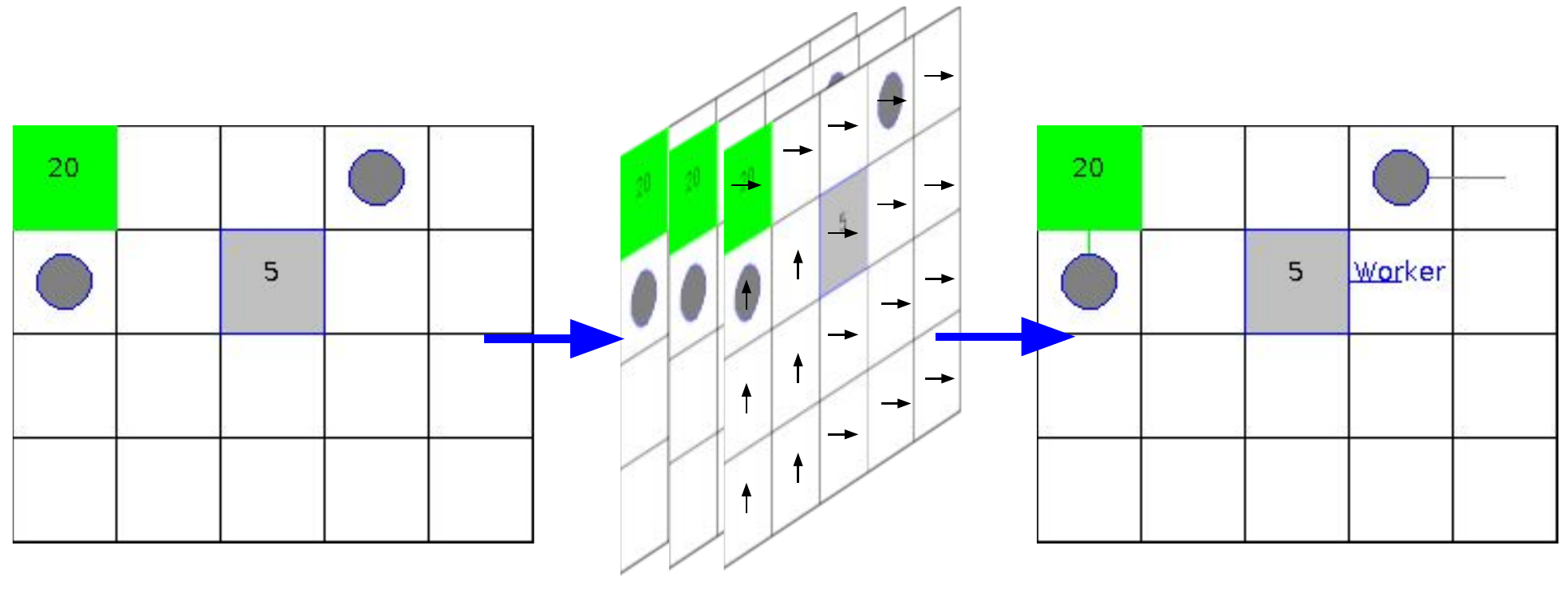}
    \caption{Gridnet predicts an action for each cell in the map (it predicts all the action component planes in one step), which then define which actions each unit will perform.}
    \label{fig:gridnet}
\end{figure}

\section{Gym-$\mu$RTS}
Gym-$\mu$RTS\footnote{\url{https://github.com/vwxyzjn/gym-microrts}} is a reinforcement learning interface for the RTS games simulator $\mu$RTS\footnote{\url{https://github.com/santiontanon/microrts}}.  Despite having a simplified implementation, $\mu$RTS captures the core challenges of RTS games, such as combinatorial action space, real-time decision-making, optionally partial observability and stochasticity. Gym-$\mu$RTS's observation space provides a series of feature maps similar to PySC2 (the StarCraft II Learning environment~\cite{vinyals2017starcraft}).
Its action space design, however, is more low-level due to its lack of AI assisted actions. In this section, we introduce their technical details.


\label{sec:details_on_murts}
\subsection{Observation Space.} 
Given a map of size $h\times w$, the observation is a tensor of shape $(h, w, n_f)$, where $n_f$ is a number of feature planes that have binary values. The observation space used in this paper uses 27 feature planes as shown in Table~\ref{tab:action-components}. The different feature planes result as the concatenation of multiple one-hot encoded features. As an example, if there is a worker player 1 with hit points equal to 1, not carrying any resources, and currently not executing any actions, then the one-hot encoding features will look like this (see Table~\ref{tab:action-components}):
\begin{align*}
    [0,1,0,0,0],  [1,0,0,0,0],  [1,0,0], \\ [0,0,0,0,1,0,0,0],  [1,0,0,0,0,0]
\end{align*}
Each feature plane contains one value for each coordinate in the map. The values for the 27 feature planes for the position in the map of such worker will thus be:
\[[0,1,0,0,0,1,0,0,0,0,1,0,0,0,0,0,0,1,0,0,0,1,0,0,0,0,0]\]

\subsection{Action Space.}
\label{sec:action-space}
Compared to traditional reinforcement learning environments, the design of the action space of RTS games is more difficult because, depending on the game state, there is a different number of units to control, and each unit might have different number of actions available. This poses a challenge for directly applying off-the-shelf DRL algorithm such as PPO that generally assume a fixed output size for the actions. Early work on RL in RTS games simply learned policies for individual units, rather than having the policy control all the units at once~\cite{jaidee2012classq}. 
To address this issue, we decompose the action space into two parts: the \emph{unit action space} (the space of possibilities for issuing actions to only one unit) and the \emph{player action space} (the space of unit actions for all the units a player owns).

\begin{table}[t]
\centering
\caption{Observation features and action components. $a_r=7$ is the maximum attack range.}
\begin{tabular}{p{2.3cm}lp{3cm}} 
\toprule
Observation Features  & Planes & Description \\
\midrule
Hit Points & 5 & 0, 1, 2, 3, $\geq 4$  \\ 
Resources & 5 & 0, 1, 2, 3, $\geq 4$  \\ 
Owner &3 & player 1, -, player 2 
\\ 
Unit Types &8 & -, resource, base, barrack, worker, light, heavy, ranged \\ 
Current Action &6& -, move, harvest, return, produce, attack\\ 
\midrule
Action Components  & Range & Description \\
\midrule
Source Unit & $[0,h \times w-1]$ & the location of the unit selected to perform an action  \\ 
Action Type & $[0,5]$ & NOOP, move, harvest, return, produce, attack  \\ 
Move Parameter & $[0,3]$ & north, east, south, west \\ 
Harvest Parameter & $[0,3]$  & north, east, south, west  \\
Return Parameter & $[0,3]$ & north, east, south, west  \\
Produce Direction Parameter & $[0,3]$ & north, east, south, west  \\
Produce Type Parameter & $[0,6]$ & resource, base, barrack, worker, light, heavy, ranged \\
Relative Attack Position & $[0,a_r^2 - 1]$  & the relative location of the unit that  will be attacked \\
\bottomrule
\end{tabular}
\label{tab:action-components}
\end{table}

In the unit action space, given a map of size $h\times w$, the unit action is an 8-dimensional vector of discrete values as specified in Table~\ref{tab:action-components}. The first component of the unit action vector represents the unit in the map to issue commands to, the second is the unit action type, and the rest of components represent the different parameters different unit action types can take. Depending on which unit action type is selected, the game engine will use the corresponding parameters to execute the action. As an example, if the RL agent issues a ``move south'' unit action to the worker at $x=3, y=2$ in a $16\times16$ map, the unit action will be encoded in the following way:
\begin{align*}
     [3+2*16,1,2,0,0,0,0,0 ]
\end{align*}
In the player action space, we compare two ways to issue player actions to a variable number of units at each frame: \textbf{Unit Action Simulation (UAS)} and \textbf{Gridnet}~\cite{han2019grid}. 

Their mechanisms are best illustrated through an example as shown in Figs.~\ref{fig:unitaActionSimulation} and~\ref{fig:gridnet}, where the player owns two workers and a base in a $4\times5$ map.


UAS calls the RL policy iteratively. At each step, the policy chooses a unit based on the \emph{source unit masks} (a vector of $h\times w$ scalars). It then chooses the action type and parameter via the \emph{unit action masks} (a vector of $6+4+4+4+4+7+49$ scalars).  We then compute a ``simulated game state'' where that action has been issued (and any potential rewards collected). Once all three units have been issued actions, the simulated game states are discarded, and the three actions are collected  and sent to the actual game environment.

Under Gridnet~\cite{han2019grid}, The RL agent receives a \emph{player action mask} (a tensor of shape $(h,w,1+6+4+4+4+4+7+49)$, where the first plane indicates if the source unit is available). It then issues actions to each cell in this map in one single step, that is, issues in total $4 * 5=20$ unit actions. The environment executes the three valid actions (actions in cells with no player-owned units are ignored).


\subsection{The Action Spaces of Gym-$\mu$RTS and PySC2}\label{subsec:action-spaces}

Although Gym-$\mu$RTS is heavily inspired by and shares many similarities with PySC2~\cite{vinyals2017starcraft}, their action space designs are considerably different. Specifically, PySC2 has designed its action space to mimic the human interface, while Gym-$\mu$RTS has a more low-level action that require actions being issued for each individual unit. This distinction is rather interesting from a research standpoint because certain challenges are easier for an AI agent and some more difficult. 

Consider the canonical task of harvesting resources and returning them to the base. In PySC2, the RL agent would need to issue two actions at two timesteps 1) select an area that has workers and 2) move the selected workers towards to a coordinate that has resources. Then, the workers will continue harvesting resources until otherwise instructed. Note that this sequence of actions is assisted by AI algorithms such as path-finding. 
After the workers harvest the resources, the engine automatically determines the closest base for returning the resources, and repeating these actions to continuously harvest resources. So the challenge for the RL agent is to learn to select the correct area and move to the correct coordinates. In Gym-$\mu$RTS, however, the RL agent can only issue primitive actions to the workers such as ``move north for one cell'' or ``harvest resource that is one cell away at north''. Therefore, it needs to constantly issue actions to control units at all times, having to learn how to perform these AI-assisted decisions from scratch\footnote{Notice, however that $\mu$RTS offers both the low-level interface and a PySC2-style interface with AI-assisted actions, but for Gym-$\mu$RTS, we only expose the former.}. 

The benefit of PySC2's approach is that it makes it easier to do imitation learning from human datasets and the resulting agent will have a fairer comparison when evaluated against humans since the AI and the human are mostly playing the same game. That being said, the human interface could be an artificial limitation to the AI system. In particular, the human interface is constructed to accommodate the human limitations: humans' eyes have limited range, so camera locations are designed to help capture larger maps, and humans have limited physical mobility, so hotkeys are set to help control a group of units with one mouse click. However, machines don't have these limitations and can observe the entire map and issue actions to all units individually.

\subsection{Reward Function}
We use a \emph{shaped reward} function to train the agents, which gives the agent +10 for winning, 0 for drawing, -10 for losing, +1 for harvesting one resource, +1 for producing one worker, +0.2 for constructing a building, +1 for each valid attack action it issues, +4 for each combat unit it produces. It gives the rewards to the frame at which the events are initialized (e.g. attack takes 5 game frames to finish, but the attack reward is given at the first frame). For reporting purposes, we also keep track of the \emph{sparse reward}, which is  +1 for winning, 0 for drawing, -1 for losing. 
The shaped reward weights are picked by hand with very little tuning.

Note this shaped reward function is similar to the one used in Open AI Five for Dota 2~\cite{Berner2019Dota2W}. Like in Open AI Five, it is possible for the agents to gain more shaped rewards by doing other good behaviors than winning the game outright. Notice we have avoided using very large win/lose rewards because anecdotally large reward numbers could cause worse performance for RL algorithms, which might be the reason why reward normalization~\cite{engstrom2019implementation} or reward clipping~\cite{mnih2013playing} have been used in previous work. 


\begin{figure*}[t]
    \centering
    \includegraphics[width=0.7\textwidth]{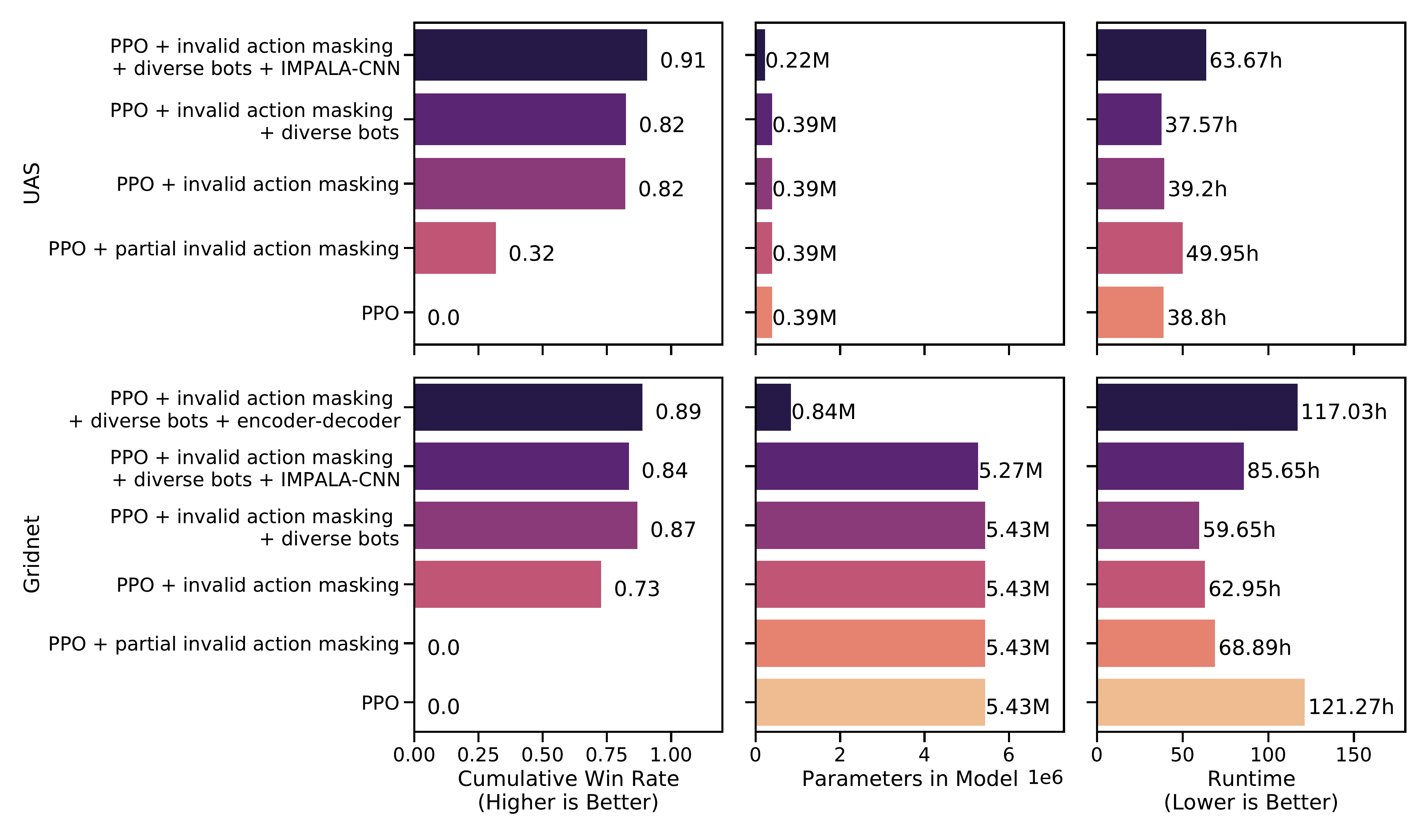}
    \caption{Ablation study for UAS and Gridnet.}
    \label{fig:ablation}
\end{figure*}


\section{Scaling DRL to Gym-$\mu$RTS}

We use PPO~\cite{schulman_2017}, a popular policy gradient algorithm, to train agents for all experiments in this paper. In addition to PPO's core algorithm, many implementation details and empirical setting also have a huge impact on the algorithm's performance~\cite{engstrom2019implementation}. 

We start with a PPO implementation that matches the implementation details and benchmarked performance in \emph{openai/baselines}~\cite{baselines}\footnote{See https://costa.sh/blog-the-32-implementation-details-of-ppo.html}, and use it along with the architecture from Mnih, et al. (denoted as Nature-CNN)~\cite{mnih2013playing} as the baseline. We train the RL agents using UAS and Gridnet by playing against CoacAI, the 2020 $\mu$RTS competition winner, in the standard \verb 16x16basesWorkers   map, where the RL agents always spawn from the top left position and end episodes after 2000 game ticks. We then incrementally include augmentations for both UAS and Gridnet and compare their relative performance. 

We run each ablation with 4 random seeds each. Then, we select the best performing seeds according to the reported sparse reward function and evaluate them 
against a pool of 11 bots with various strategies that have participated in previous $\mu$RTS competitions (other competition bots are not included due to either staleness or difficulty to set up) and 2 baseline bots which are mainly used for testing. 
All $\mu$RTS bots are configured to use their $\mu$RTS competition parameters and setups. The name, category and best result of these bots are listed in Table~\ref{tab:bots}. The evaluation involves playing 100 games against each bot in the pool for 4000 maximum game ticks, and we report the cumulative win rate, the model size, and total run time in Figure~\ref{fig:ablation}.
To further provide insights, we record videos of the RL agents against each of the bots in the pool and make them publicly available\footnote{
\url{https://wandb.ai/vwxyzjn/gym-microrts-paper-eval/reports/Final-Eval--Vmlldzo0OTY1Mzc}
}. Let us now describe the different augmentations we added on top of PPO.



\subsection{Action Composition}

After having solved the problem of issuing actions to a variable number of units (via either UAS or Gridnet), the next problem is that even the action space of a single unit is too large. Specifically, to issue a single action $a_t$ in $\mu$RTS using UAS, according to Table~\ref{tab:action-components}, we have to select a Source Unit, Action Type, and its corresponding action parameters. So in total, there are $hw\times6\times4\times4\times4\times4\times6\times a_r^2 = 9216(hwa_r^2) $ number of possible discrete actions, which includes many invalid actions, which is huge even for small maps (about 50 million in the map size we use in this paper).

To address this problem, we use {\em action composition}, 
where we consider an action as composed of some smaller \emph{independent} discrete actions. Namely, $a_t$ is composed of a set of smaller actions 
$D = \{a_{t}^{\text{Source Unit}}$, $a_{t}^{\text{Action Type}}$, $a_{t}^{\text{Move Parameter}}$, $a_{t}^{\text{Harvest Parameter}}$, $ a_{t}^{\text{Return Parameter}}$, $a_{t}^{\text{Produce Direction Parameter}}$, $ a_{t}^{\text{Produce Type Parameter}}$, $a_{t}^{\text{Relative Attack Position}} \}$.
And the policy gradient is updated in the following way (without considering the PPO's clipping for simplicity):
\begin{align*}
         \sum_{t=0}^{T-1}\nabla_{\theta}\log\pi_{\theta}(a_t|s_t)G_t  &= \sum_{t=0}^{T-1}\nabla_{\theta}  \left( \sum_{a^{d}_{t}\in D} \log\pi_{\theta}(a^{d}_{t}|s_t) \right)G_t\\
         &= \sum_{t=0}^{T-1}\nabla_{\theta}  \log \left( \prod_{a^{d}_{t}\in D} \pi_{\theta}(a^{d}_{t}|s_t) \right)G_t
\end{align*}
Implementation-wise, for each action component, the logits of the corresponding shape are output by the policy, which we refer to as action component logits. Each action $a^{d}_{t}$ is sampled from a softmax distribution parameterized by these action component logits. In this way, the algorithm has to generate $hw+6+4+4+4+4+6+a_r^2 = hw + 36 + a_r^2$ logits, significantly less than $9216(hwa_r^2)$ (301 vs 50 million). 


\subsection{Invalid Action Masking}

The next most important augmentation in our experiments is {\em invalid action masking}, which ``masks out'' invalid actions out of the action space (by exploiting the fact that we know the rules of the game), significantly reducing it. This is used in PySC2~\cite{vinyals2017starcraft}, OpenAI Five~\cite{Berner2019Dota2W}, and a number of related work with large action spaces~\cite{samvelyan2019starcraft}.

\begin{figure}[t]
     \centering
     \begin{subfigure}[b]{0.35\textwidth}
         \centering
         \includegraphics[width=\textwidth]{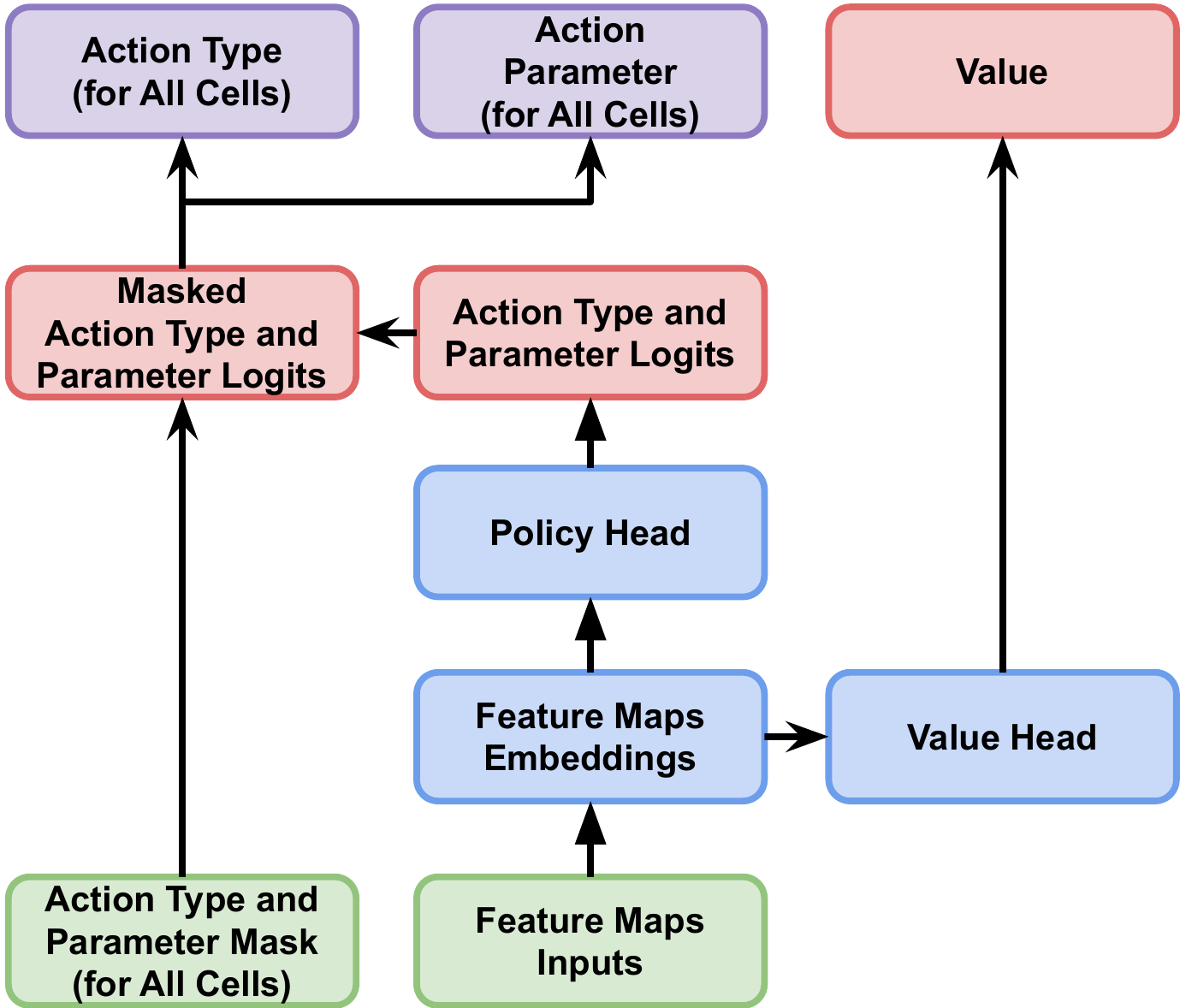}
         \caption{Gridnet.}
         \label{fig:gridnet-architecture}
     \end{subfigure}
     \hfill
     \begin{subfigure}[b]{0.48\textwidth}
         \centering
         \includegraphics[width=\textwidth]{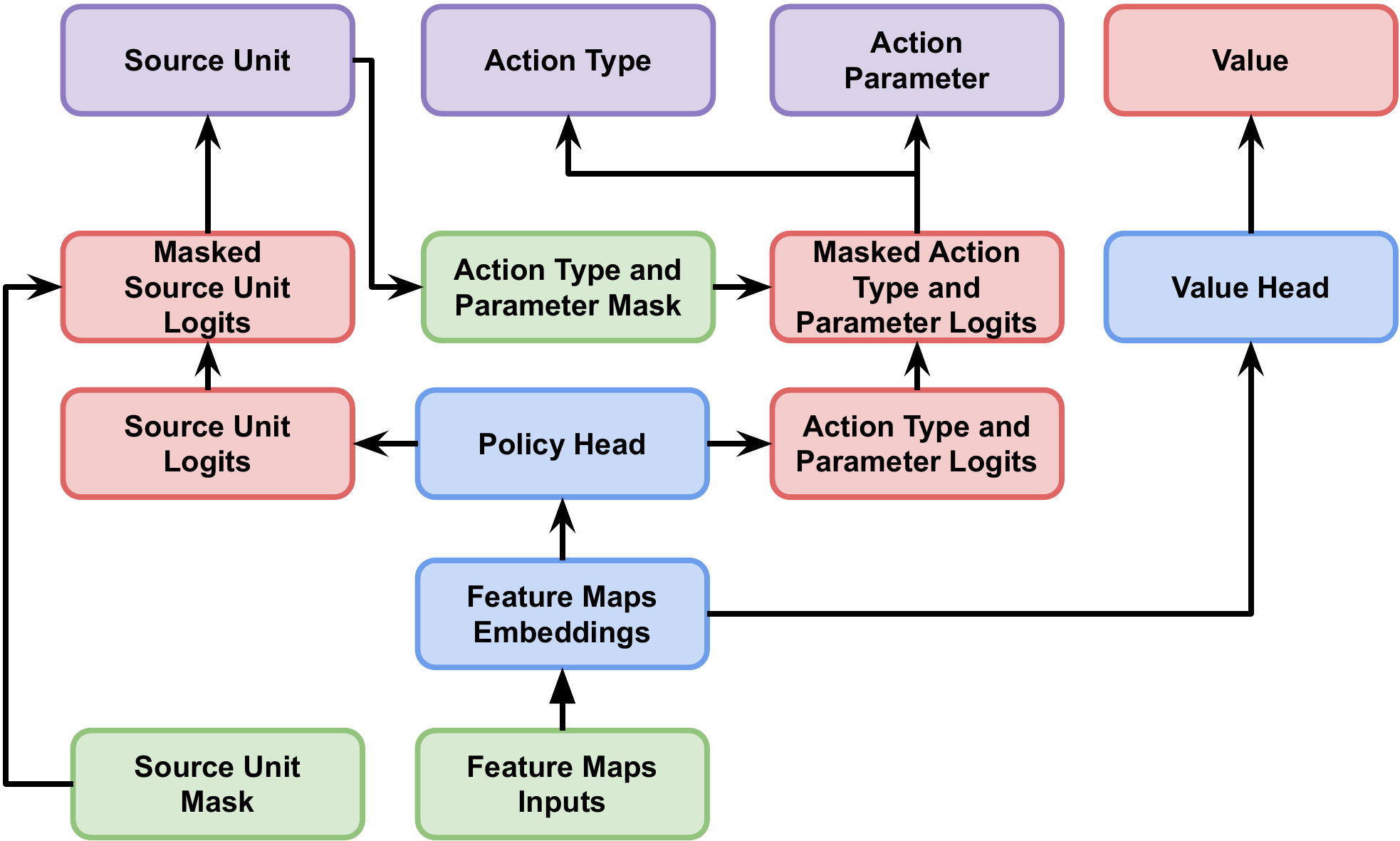}
         \caption{UAS.}
         \label{fig:uas-architecture}
     \end{subfigure}
        \caption{Neural network architectures for Gridnet and UAS. The green boxes are (conditional) inputs from the environments, blue boxes are neural networks, red boxes are outputs, and purple boxes are sampled outputs.}
        \label{fig:architecture}
\end{figure}



Masks are generated and being applied as shown in Figure~\ref{fig:architecture}. Under UAS, the agent would first sample a source unit based on the source units masks of shape $(hw)$, then query the game client for the action type and parameter mask of the said units with shape $(78)$. Under Gridnet, the agent would receive all the masks up front on source unit, action type and parameter with shape $(hw, 79)$, where the first plane of 79 is the mask on the source unit selection. Note that in both cases, the agent received a \emph{full} action mask that in a sense significantly reduce the search space. In contrast, PySC2 and SMAC (the StarCraft Multi-Agent Challenge)~\cite{samvelyan2019starcraft} would only provide a \emph{partial} mask on the action type, and the logits of action parameters are unmasked (our action types and action parameters are function identifiers and arguments in PySC2's term). This could explain why invalid action masking does not seem to cause as drastic of a difference in PySC2 as shown by Kanervisto et al.~\cite{kanervisto2020action}.

In the interest of ablation study, we also conduct experiments that provide masking on the action types but not the action parameters, which is more similar to PySC2's settings. As shown in Figure~\ref{fig:ablation}, we see that having only a partial mask has little impact whereas having the full mask considerably improves performance. Although the action space and PySC2 is quite different as discussed above, masking all invalid actions maximally reduces the action space, hence simplifying the learning task. We therefore believe that the PySC2 agents could receive a performance boost by providing masks on function arguments as well.

\subsection{Other augmentations}
This section details other additional augmentations that contribute to the agents' performance, but not as much as the previous two (which are essential for having an agent that even starts learning to play the full game).

\subsubsection{Diverse Opponents}
The  baseline setting is to train the agents against CoacAI. However, this lacks a diversified experience and when evaluating, we frequently see the agents being defeated by AIs as simple as WorkerRush. To help alleviate this problem, we train the agents against a diverse set of built-in bots. Since we train with 24 parallel environments for PPO, we set 18 of these environments to have CoacAI as the opponent, 2 to have RandomBiasedAI, 2 to have WorkerRush, and 2 to have LightRush. Per Figure~\ref{fig:ablation}, we see a rather significant performance boost for Gridnet, whereas in UAS the performance boost is milder.

\subsubsection{Nature-CNN vs Impala-CNN vs Encoder-Decoder}
To seek better neural network architectures, we experimented with the use of residual blocks~\cite{He_2016} (denoted IMPALA-CNN), which have been shown to improve the agents' performance in several domains like DMLab~\cite{espeholt2018impala}. 
Additionally, Han et al. ~\cite{han2019grid} also experimented with an encoder-decoder network in Gridnet, so we also conducted experiments using this architecture. Per the ablation study in Figure~\ref{fig:ablation}, we see IMPALA-CNN helps with the performance of UAS whereas encoder-decoder benefits Gridnet.

\begin{table}[t]
\centering
\caption{The previous $\mu$RTS competition bots.}
    \begin{tabular}{lll}
        \toprule
        Name & Category & Best result \\
        \midrule
        CoacAI & Scripted & 1st place in 2020 \\
        Tiamat & MCTS-based & 1st place in 2018 \\
        MixedBot & MCTS-based & 2nd place in 2019 \\
        Droplet & MCTS-based & 3rd place in 2019 \\
        Izanagi & MCTS-based & 4th place in 2019 \\
        Rojo & MCTS-based & 5th place in 2020 \\
        LightRush & Scripted & 6th place in 2020 \\
        GuidedRojoA3N & MCTS-based & 7th place in 2020 \\
        WorkerRush & Scripted & 8th place in 2020 \\
        NaiveMCTS & MCTS-based & 9th place in 2020 \\
        RandomBiasedAI & Scripted & 10th place in 2020 \\
        Random & Scripted & - \\
        PassiveAI & Scripted & - \\
        \bottomrule
    \end{tabular}
\label{tab:bots}
\end{table}

\begin{figure}[t]
    \centering
    \includegraphics[width=0.9\columnwidth]{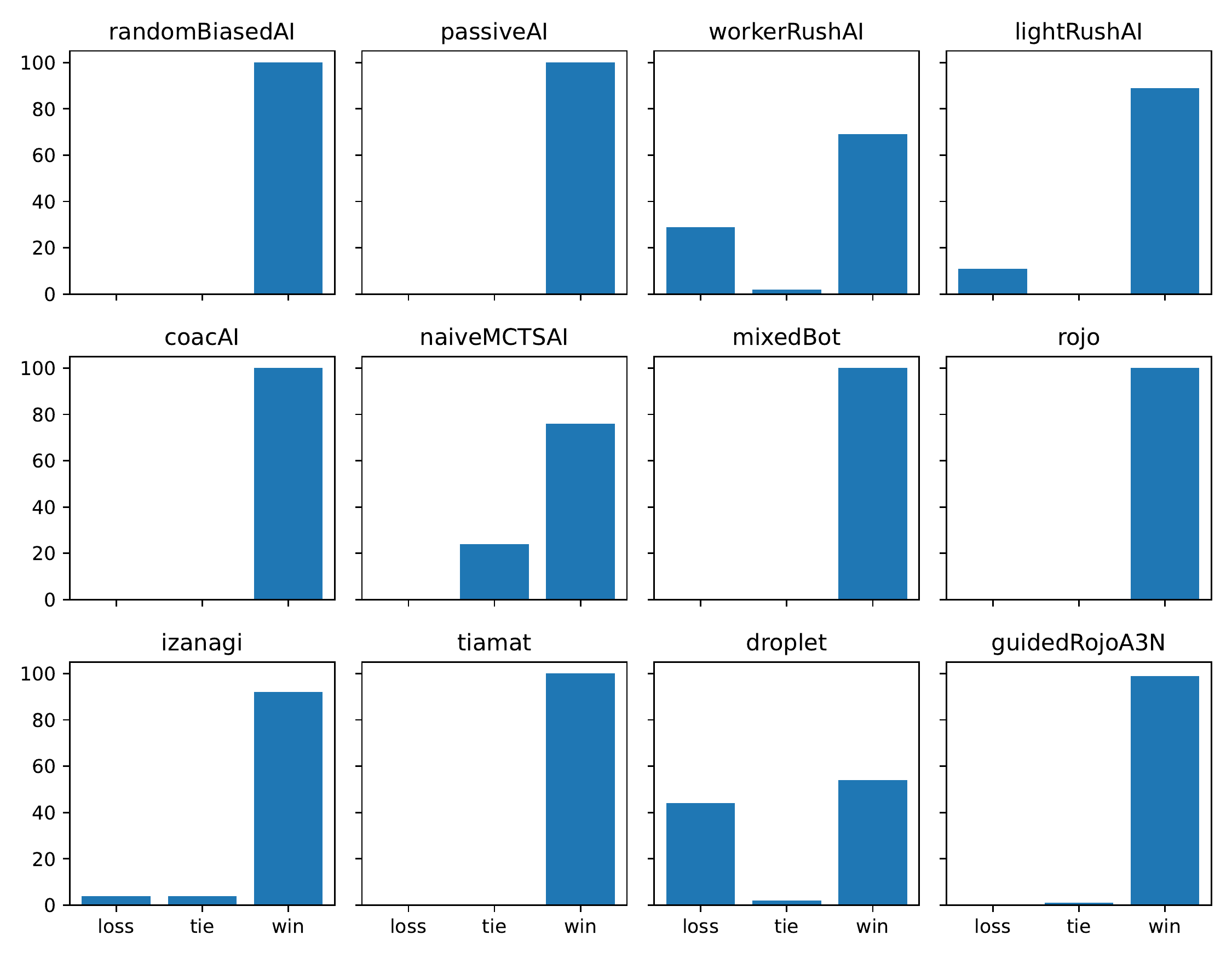}
    \caption{Match results: the y-axis shows the number of losses, ties, and wins against AIs listed in Table~\ref{tab:bots}. The Random bot's match result is excluded for presentation purposes.}
    \label{fig:matches}
\end{figure}

\begin{figure}[t]
    \centering
    \includegraphics[width=0.9\columnwidth]{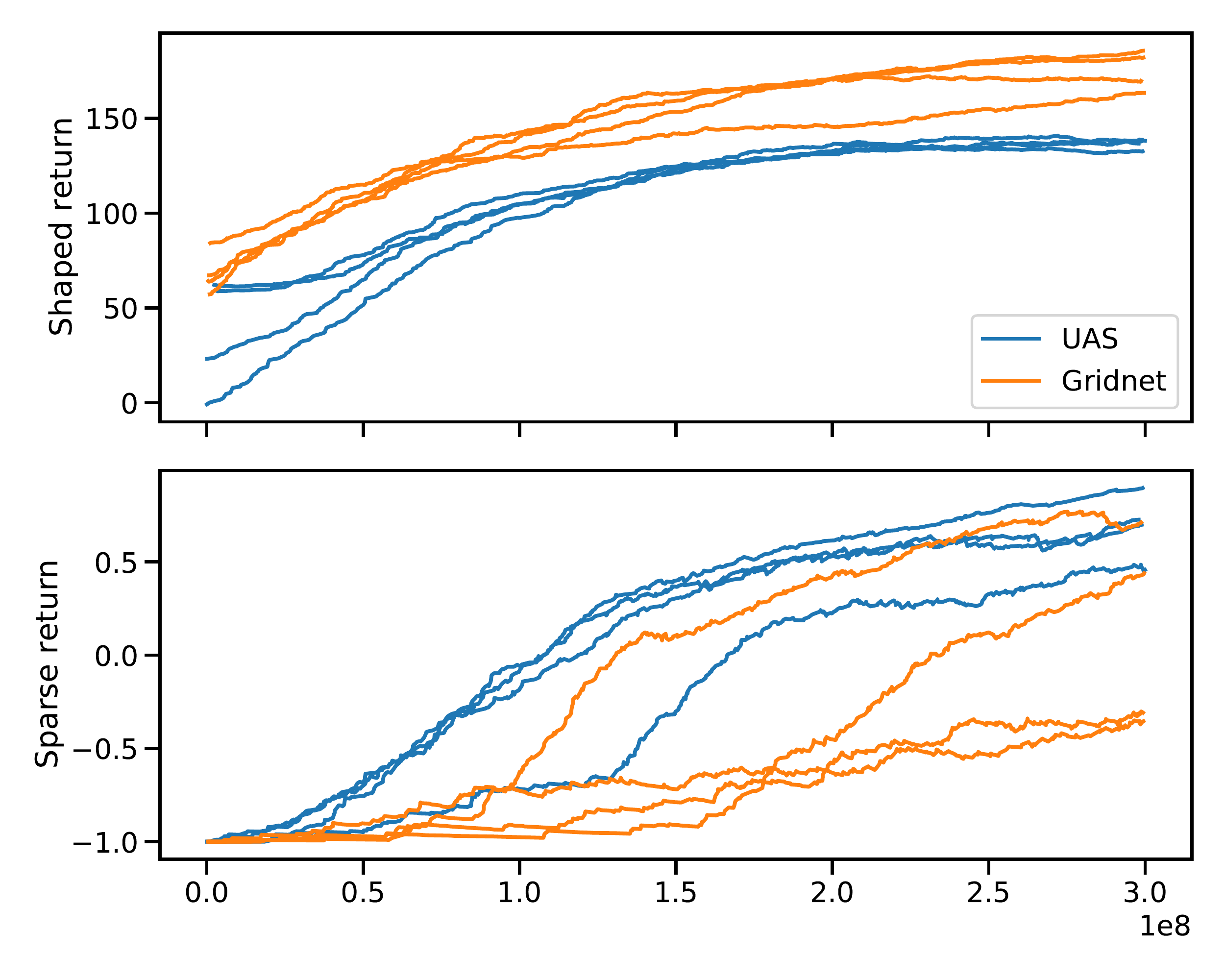}
    \caption{The shaped and sparse return over training steps for all 4 random seeds of \emph{PPO + invalid action masking} for Gridnet and UAS. The curve is smoothed using exponential moving average with weight 0.99.}
    \label{fig:ppo-gridnet-uas}
\end{figure}

\section{Discussion}

{\bf Establishing a SOTA in Gym-$\mu$RTS.} 
According to Figure~\ref{fig:ablation}, our best agent consists of \emph{ppo + coacai + invalid action masking + diverse opponents + impala cnn}, reaching the cumulative win rate of 91\%. Additionally, Figure~\ref{fig:matches} shows the specific match results, showing this agent can outperform all other bots in the pool. Note that in the  $\mu$RTS competition settings the players could start in two different locations of the map whereas our agent always start from the top left. Nevertheless, due to the symmetric nature of the map, we could address this issue by ``rotating'' the map when needed so that both starting locations look the same to our agent. Therefore, our agent establishes the state of the art for $\mu$RTS in the \verb 16x16basesWorkers  map. Note that generalizing to handle a variety of maps (including the asymmetric ones) in $\mu$RTS competition settings is part of our future work (also note that some work on StarCraft II also focused in the one map setting~\cite{han2020tstarbot}, while still requiring large computation budgets).

Our best agent struggles the most against the Droplet bot, which typically uses a worker rush strategy, but enhanced thanks to MCTS search. Droplet usually defeats our agent by destroying the first barracks our agent makes, which is a rare experience with other bots. As a result, our agent would keep trying to build a barracks until it exhausts its resources, at which point, Droplet would have more left over resources, build more workers and eventually defeat our agent. However, if by chance our agent successfully builds and protects the first barracks and combat units, it is usually able to defeat Droplet. As part of our future work, we would like to include agents like Droplet in our training process. However, search-based bots like Droplet significantly decrease the speed of training. 

{\bf Hardware Usage and Training Time.}  Most of our experiments are conducted using 3 vCPUs, 1 GPU, and 16GB RAM. According to Figure~\ref{fig:ablation}, the experiments take anywhere from 37 hours to 117 hours, where our SOTA agent takes 63 hours. 


{\bf Model size vs performance.} Overall, Gridnet models have more parameters compared to the UAS models. This is because Gridnet  predicts the action type and parameter logits for every cell in the map. 
We did not find a strong correlation between the model's size (in number of trainable parameters) and the performance of the agents. As shown in Figure~\ref{fig:ablation}, it is clear that the techniques such as invalid action masking or different neural network architectures are more important to the performance than the sheer number of the model's trainable parameters in our experiments.

{\bf Variance w.r.t. Shaped and Sparse Reward.} 
In almost all experiments conducted in this paper, we observe the RL agents are able to optimize against the shaped rewards well, showing little variance across different random seeds; however, this is not the case with respect to the sparse reward (win/loss). We report the sum of shaped rewards and sparse rewards in the episode as \emph{shaped return} and \emph{sparse return} respectively  in Figure~\ref{fig:ppo-gridnet-uas}, where
we usually see little difference in the shaped return when the sparse (win/loss) return could be drastically different. This is a common drawback with reward shaping: agents sometimes overfit to the shaped rewards instead of sparse rewards.


{\bf UAS vs Gridnet.} 
Figure~\ref{fig:ppo-gridnet-uas} shows  a typical result where Gridnet is able to get much higher shaped return, but it receives relatively similar sparse return as UAS. Upon further inspection of the agents actual behaviors, we found the Gridnet agents obtain higher shaped return by 1) producing more barracks, 2) producing more combat units, and 3) harvesting more resources effectively. In fact, Gridnet agents learn to harvest resources using three workers, which is a behavior we haven't observed in any existing bots. We suspect this difference is due to how rewards are attributed in UAS vs Gridnet. UAS attributes rewards to unit actions \emph{individually}, while Gridnet attributes the rewards to the player action \emph{collectively}. 

Depending on the implementation, Gridnet agents usually have many more trainable parameters. Also, when the player owns a relatively small amount of units, it is faster to step the environment using UAS because Gridnet has to predict an action for all the cells in the map; however, when the player owns a large number of units, Gridnet's mechanism becomes faster because UAS has to do more simulated steps and thus more inferences.




{\bf The Amount of Human Knowledge Injected.} 
In our best trained agents, there are usually three sources of human knowledge injected: 1) the reward function, 2) invalid action masking, and 3) the use of human-designed bots such as CoacAI. In comparison, AlphaStar uses 1) human replays, 2) its related use of Statistics $z$ and Supervised KL divergence~\cite{vinyals_2019}, and 3) invalid action masking. 


\section{Conclusions and Future Work}

We present a new efficient library, Gym-$\mu$RTS, which allows DRL research to be realized in the complex RTS environment $\mu$RTS. Through Gym-$\mu$RTS, we conducted ablation studies on techniques such as action composition, invalid action masking, diversified training opponents, and novel neural network architectures, providing insights on their importance to scale agents to play the full game of $\mu$RTS.
Our agents can be trained on a single CPU+GPU within 2-4 days, which is a reasonable hardware and time budget that is available to many researchers outside of large research labs



For future work, we would like to consider multiple maps and the partial observability setting of $\mu$RTS (i.e. fog-of-war). Additionally, we also want to experiment with selfplay, which further reduces human knowledge injected such as the human-designed bots we used in this paper.

\section*{Acknowledgments}
 This research utilized Queen Mary's Apocrita HPC facility, supported by QMUL Research-IT~\cite{king_thomas_2017_438045}.

\bibliographystyle{IEEEtran}
\bibliography{references}

\clearpage
\onecolumn
\appendix
\addcontentsline{toc}{section}{Appendices}
\renewcommand{\thesubsection}{\Alph{subsection}}

\subsection{Details on the Training Algorithm Proximal Policy Optimization}
\label{sec:details_on_ppo}
The DRL algorithm that we used to train the agent is Proximal Policy Optimization (PPO)~\cite{schulman_2017}, one of the state-of-the-art algorithms available. The hyper-parameters of our experiments can be found in Table~\ref{tab:params}. 
In addition to the techniques of action composition and invalid action masking, whose details are highlighted in the main text, our PPO implementation matches the implementation details in \emph{openai/baselines}. Here we enumerate these details, including a footnote directing the readers to the original files in the  \emph{openai/baselines}~\cite{baselines} that implements these details.

\begin{table}[h]
\centering
\caption{The list of hyperparameters and their values.}
\begin{tabular}{ll} 
\toprule
Parameter Names  & Parameter Values\\
\midrule
$N_{total}$ Total Time Steps & 300,000,000  \\ 
$N_{mb}$ Number of Mini-batches & 4 \\
$N_{envs}$ Number of Environments & 24 \\
$N_{steps}$ Number of Steps per Environment & 256 for UAS and 512 for Gridnet \\
$\gamma$ (Discount Factor) & 0.99 \\ 
$\lambda$ (for GAE) & 0.95 \\ 
$\varepsilon$ (PPO's Clipping Coefficient) & 0.1 \\ 
$\omega$ (Gradient Norm Threshold)& 0.5 \\
$K$ (Number of PPO Update Iteration Per Epoch)& 4 \\
$\alpha$ Learning Rate &  0.00025 Linearly Decreased to 0 \\
& over the Total Time Steps\\
$c1$ (Value Function Coefficient)& 0.5\\
$c2$ (Entropy Coefficient)& 0.01\\
$N_{updates}$ (Total Number of Updates) & $N_{total}/(N_{mb}N_{envs})$\\
\bottomrule
\end{tabular}
\label{tab:params}
\end{table}  


\begin{enumerate}
    \item \textbf{Clipped Surrogate Objective\footnote{\url{https://github.com/openai/baselines/blob/ea25b9e8b234e6ee1bca43083f8f3cf974143998/baselines/ppo2/model.py\#L81-L86}}:} This core feature of PPO.
    \item \textbf{Generalized Advantage Estimation (GAE)\footnote{\url{https://github.com/openai/baselines/blob/ea25b9e8b234e6ee1bca43083f8f3cf974143998/baselines/ppo2/runner.py\#L56-L65}}:} Although Schulman, et al.~\cite{schulman_2017} just use the abstraction of advantage estimate in the PPO's objective, the implementation in \emph{openai/baselines} specifically use GAE.
    \item \textbf{Normalization of Advantages\footnote{\url{https://github.com/openai/baselines/blob/ea25b9e8b234e6ee1bca43083f8f3cf974143998/baselines/ppo2/model.py\#L139}}:} After calculating the advantages based on GAE, the advantages vector is normalized by subtracting its mean and divided by its standard deviation.
    \item \textbf{Value Function Loss Clipping\footnote{\url{https://github.com/openai/baselines/blob/ea25b9e8b234e6ee1bca43083f8f3cf974143998/baselines/ppo2/model.py\#L68-L75}}:} The PPO implementation of {\em  openai/baselines} clips the value function loss in a manner that is similar to the PPO's clipped surrogate objective:
    \[V_{loss} =\max \left[\left(V_{\theta_{t}}-V_{t a r g}\right)^{2},\left(V_{\theta_{t-1}} + \mbox{clip}\left(V_{\theta_{t}}-V_{\theta_{t-1}}, -\varepsilon, \varepsilon\right)\right)^{2}\right]\]
    where $V_{t a r g}$ is calculated by adding $V_{\theta_{t-1}}$ and the  $A$ calculated by General Advantage Estimation\cite{schulman2015high}.
    \item \textbf{Overall Loss Includes Entropy Loss\footnote{\url{https://github.com/openai/baselines/blob/ea25b9e8b234e6ee1bca43083f8f3cf974143998/baselines/ppo2/model.py\#L91}}:} The overall loss is constructed with the policy gradient loss, value loss, and entropy loss. The entropy loss intuitively encourages the exploration by encouraging the action probability distribution to be more chaotic. 
    \item \textbf{Adam Learning Rate Annealing\footnote{\url{https://github.com/openai/baselines/blob/ea25b9e8b234e6ee1bca43083f8f3cf974143998/baselines/ppo2/ppo2.py\#L135}}:} The Adam~\cite{kingma2014adam} optimizer's learning rate is set to decay as the number of timesteps agent trained increase.
    \item \textbf{Mini-batch updates\footnote{\url{https://github.com/openai/baselines/blob/ea25b9e8b234e6ee1bca43083f8f3cf974143998/baselines/ppo2/ppo2.py\#L160-L162}}:} The PPO implementation of the {\em  openai/baselines} also uses mini-batches to compute the gradient and update the policy instead of the whole batch data such as in {\em open/spinningup}.
    The mini-batch sampling scheme, however, still makes sure that every transition is sampled only once, and that the all the transitions sampled are actually for the network update. 
    \item \textbf{Global Gradient Clipping\footnote{\url{https://github.com/openai/baselines/blob/ea25b9e8b234e6ee1bca43083f8f3cf974143998/baselines/ppo2/model.py\#L107}}:} For each update iteration in an epoch, the gradients of the policy and value network are clipped so that the ``global $\ell_{2}$ norm'' (i.e. the norm of the concatenated gradients of all parameters) does not exceed 0.5.
    \item \textbf{Orthogonal Initialization  of weights\footnote{\url{https://github.com/openai/baselines/blob/ea25b9e8b234e6ee1bca43083f8f3cf974143998/baselines/a2c/utils.py\#L58}}:} The weights and biases of fully connected layers use with orthogonal initialization scheme with different scaling. 
    \item \textbf{Global Gradient Clipping\footnote{\url{https://github.com/openai/baselines/blob/ea25b9e8b234e6ee1bca43083f8f3cf974143998/baselines/ppo2/model.py\#L107}}:} For each update iteration in an epoch, the gradients of the policy and value network are clipped so that the ``global $\ell_{2}$ norm'' (i.e. the norm of the concatenated gradients of all parameters) does not exceed 0.5.
    \item \textbf{The Use of Parallel Environment\footnote{\url{https://github.com/openai/baselines/blob/ea25b9e8b234e6ee1bca43083f8f3cf974143998/baselines/common/cmd_util.py\#L22}}:} PPO uses parallel environments to speed up execution. In these environments, PPO collects ``fixed-length trajectory segments'' as rollouts (e.g. after the ``fixed-length trajectory segments'' of the episode, the observation in the current segment becomes the first observation in the next segment; if the episode has terminated, replace the terminal observation with the observation from the reset environments).
    \item \textbf{The Epsilon Parameter of Adam Optimizer\footnote{\url{https://github.com/openai/baselines/blob/ea25b9e8b234e6ee1bca43083f8f3cf974143998/baselines/ppo2/model.py\#L100}}:} By default, this parameter is not in the list of configurable hyper-parameters of PPO, but it was set to $10^{-5}$.
    \item \textbf{Sharing Hidden Layers for Policy and Value Functions\footnote{\url{https://github.com/openai/baselines/blob/ea25b9e8b234e6ee1bca43083f8f3cf974143998/baselines/common/policies.py\#L157}}:} The hidden layers for the policy and value functions share the same weights and biases. Although I don't think it makes much difference for the performance, the computational costs are definitely reduced since only one set of hidden layers needs to be optimized.
\end{enumerate}

\subsection{Invalid Action Masking Implementation}
The most important augmentation in our experiments is {\em invalid action masking}. However, this technique introduces subtle issues when implementation is involved: existing literature usually does not describe the specific details, whereas in open source repositories we see different implementations. Hence, we describe a canonical implementation and discuss a variant.


Consider this example: suppose we are generating a unit action in Gym-$\mu$RTS at state $s_0$, and we have already sampled a valid source unit (worker 1) and an action type (move). Imagine now that the network outputs the following logits for the move parameter: $l=[1.0,1.0,1.0,1.0]$, resulting in a probability distribution (via softmax) of: $\pi_{\theta}(\cdot \vert s_0) = \text{softmax}(l) = [0.25, 0.25, 0.25, 0.25]$

At this point, regular policy gradient algorithms will sample an action from $\pi_{\theta}(\cdot \vert s_0)$. Suppose $a_0$ is sampled from $\pi_{\theta}(\cdot \vert s_0)$, and the policy gradient is calculated as follows:
\begin{align*}
    g_{\text{policy}} &=  \mathbb{E}_{\tau}\left [\nabla_{\theta} \sum_{t=0}^{T-1} \log\pi_{\theta}(a_t|s_t)G_t \right] \\
    &=\nabla_{\theta}\log\pi_{\theta}(a_0|s_0)G_0 \\
    &= [ 0.75, -0.25, -0.25, -0.25]\\
    \text{where }    (\nabla_{\theta}\log \text{softmax}(\theta)_j)_i &= \begin{cases}
 (1 -  \frac{\exp(l_j)}{\sum_j \exp(l_j)}) &\text{if $i=j$}\\
 \frac{-\exp(l_j)}{\sum_j \exp(l_j)} &\text{otherwise}
\end{cases}
\end{align*}
Now suppose $a_2$ is invalid for state $s_0$, and the only valid actions are $a_0, a_1, a_3$. The canonical implementation is to ``mask out'' the logits corresponding to the invalid actions\footnote{See  \url{https://bit.ly/3cnMOAb} for the original source code.}.
That is, it replaces the logits of the invalid actions by a large negative number $M$ (e.g. $M = -1 \times 10^8$).  Unless otherwise specified, this is the implementation we used for experiments.
Let us use $\text{inv}_{s}$ to denote this masking process, and we can calculate the re-normalized probability distribution $\pi'_{\theta}(\cdot \vert s_0)$ as the following:
\begin{align*}
    &\pi'_{\theta}(\cdot \vert s_0)= \text{softmax}(\text{inv}_{s}([l_0, l_1, l_2, l_3]))
    \\
    &=\text{softmax}([l_0, l_1, M, l_3]) = [\pi'_{\theta}(a_0|s_0), \pi'_{\theta}(a_1|s_0), \epsilon, \pi'_{\theta}(a_3|s_0)] 
    \\
    &\approx [0.33, 0.33, 0.00, 0.33] \nonumber
\end{align*}
where $\epsilon$ is the resulting probability of the masked invalid action, which should be a small number. If $M$ is chosen to be sufficiently negative, the probability of choosing the masked invalid action $a_2$ will be virtually zero. After finishing the episode, the policy is updated according to the following gradient, which we refer to as the \emph{invalid action policy gradient}.
\begin{align*}
    g_{\text{invalid action policy}} &= \mathbb{E}_{\tau}\left [\nabla_{\theta} \sum_{t=0}^{T-1} \log\pi'_{\theta}(a_t|s_t)G_t \right] \\
    &=\nabla_{\theta}\log\pi'_{\theta}(a_0|s_0)G_0  \\
    &\approx [ 0.67, -0.33,  0.00, -0.33]\nonumber
\end{align*}
Hence, the canonical implementation both ``renormalizes the probability distribution'' and also makes the gradient corresponding to the logits of the invalid action to zero.

Another implementation of invalid action masking is to still sample actions using the masks, but update the gradient as if the masks had never been applied, as done in~\cite{samvelyan2019starcraft}. So using the example above, the gradient will look exactly the same as $g_{\text{policy}}$, but we refer to this implementation as \emph{naive invalid action masking} since the behavior policy does not match the target policy. Although it is shown naive invalid action masking can perform just as well as the canonical counterparts in simple tasks~\cite{huang2020closer}, Figure~\ref{fig:full-ablation} provides clear evidence that the canonical implementation is superior in larger tasks such as the full-game.

\subsection{Selfplay Experiments}
\label{sec:selfplay-experiments}
Another augmentation we tried is selfplay, which is a crucial component in recent work such as AlphaStar~\cite{vinyals_2019} and OpenAI Five~\cite{Berner2019Dota2W}. 
If the agents issue actions via Gridnet, selfplay can be implemented naturally with the parallel environments of PPO. That is, assume there are 2 parallel environments, we can spawn 1 game under the hood and use return player 1 and 2's observation for the first and second parallel environments, respectively and take the player actions respectively. However, note that the agents in the selfplay experiments are learning to handle both starting locations of the map, which is a different setting for other experiments, and it is for this reason we did not include the selfplay experiments in the main text. 

\begin{figure*}[t]
    \centering
    \includegraphics[width=\columnwidth]{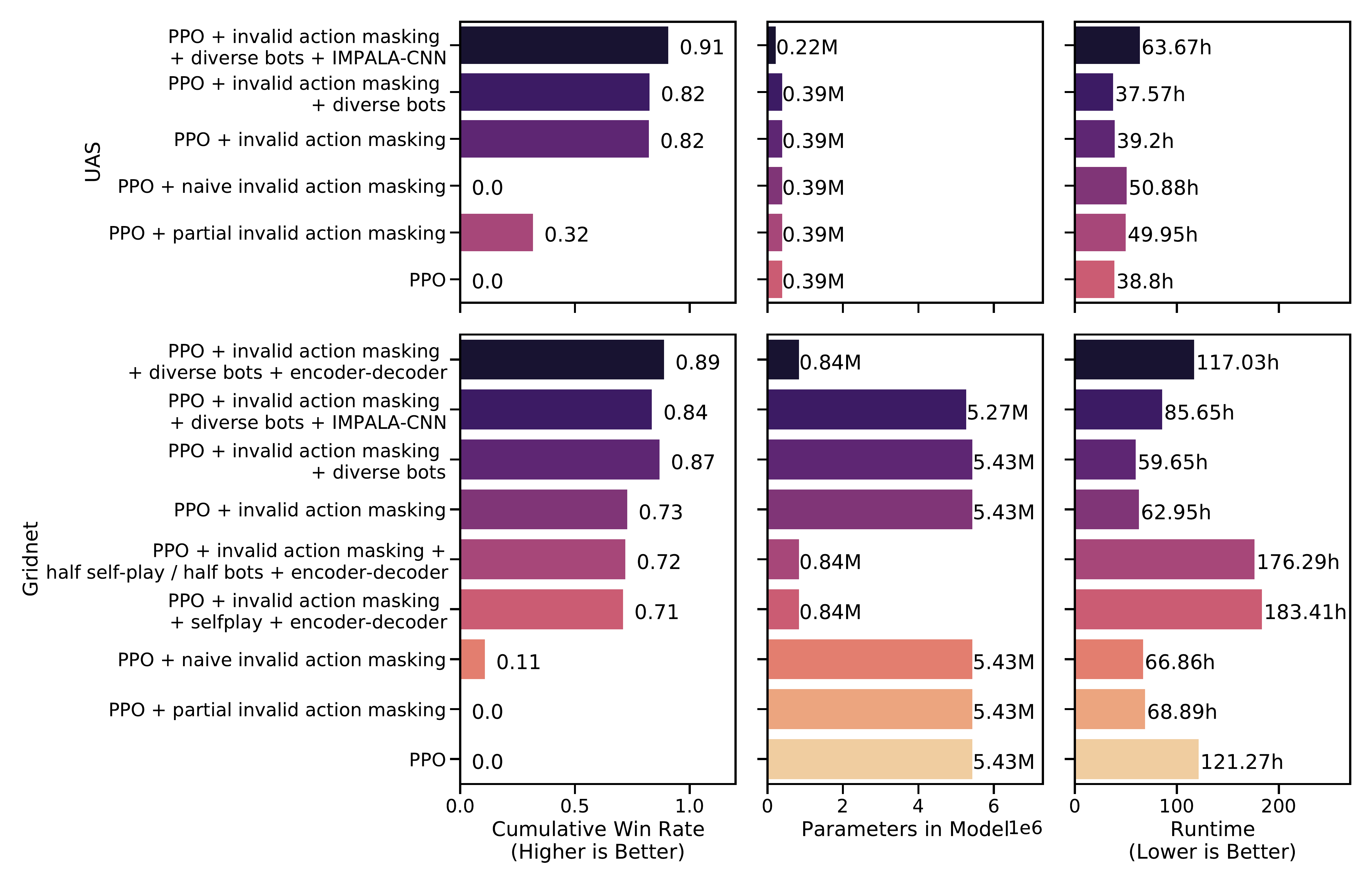}
    \caption{Full ablation study for UAS and Gridnet.}
    \label{fig:full-ablation}
\end{figure*}

We examine two settings:  1) \emph{selfplay} with 12 underlying selfplay games for 24 parallel environments, and 2) \emph{half selfplay / half bots} with 8 underlying selfplay games for 16 parallel environments and 8 bot parallel environments (2 CoacAI, 2 RandomBiasedAI, 2 LightRush, and 2 WorkerRush). Additionally, we always use the same network to predict actions for these 24 parallel environments, which is a simpler setting (i.e. always playing against the latest self) compared to the more sophisticated selfplay settings in AlphaStar~\cite{vinyals_2019} and OpenAI Five~\cite{Berner2019Dota2W}.  When the agents are evaluated using the same setting as other experiments (i.e. spawn from the top right), Figure~\ref{fig:full-ablation} shows both settings do not perform particularly well. This likely suggests a more sophisticated method for selfplay such as league training~\cite{vinyals_2019} is required to create strong agents. But again, notice the selfplay agents learn to play the game with both starting locations, so they do not compare fairly with the other experiments that always start from the top left of the map.

\subsection{Infrastructure}
Although the cost of training one instance of agent in Gym-$\mu$RTS is relatively cheap (1 GPU, 4 CPU for 3-4 days), conducting the ablation studies with 4 random seeds is less so (about 6000 CPU and GPU hours in total). In practice, we leverage Amazon Web Services (AWS) as an easily accessible cloud provider to help scale the experiments. 
Specifically, we package our codebase into a docker container and submit experiments to AWS Batch, which is configured to use the g4dn.xlarge instances (4 vCPU 16 GiB RAM, 1 GPU) to process jobs. Our training script periodically saves the model to Weights and Biases; this allows us to reduce computational costs by using the Spot instances that can be preempted at any time. When the instances get preempted, we simply submit another job to resume the run by loading the saved model from Weights and Biases. That said, some runs like the encoder-decoder experiments are conducted using Queen Mary University's computing cluster.

\subsection{Learning curves and match results}
\label{sec:lc-mr}

\begin{figure}[h]
    \begin{subfigure}[t]{0.5\hsize}
        \includegraphics[width=0.9\linewidth]{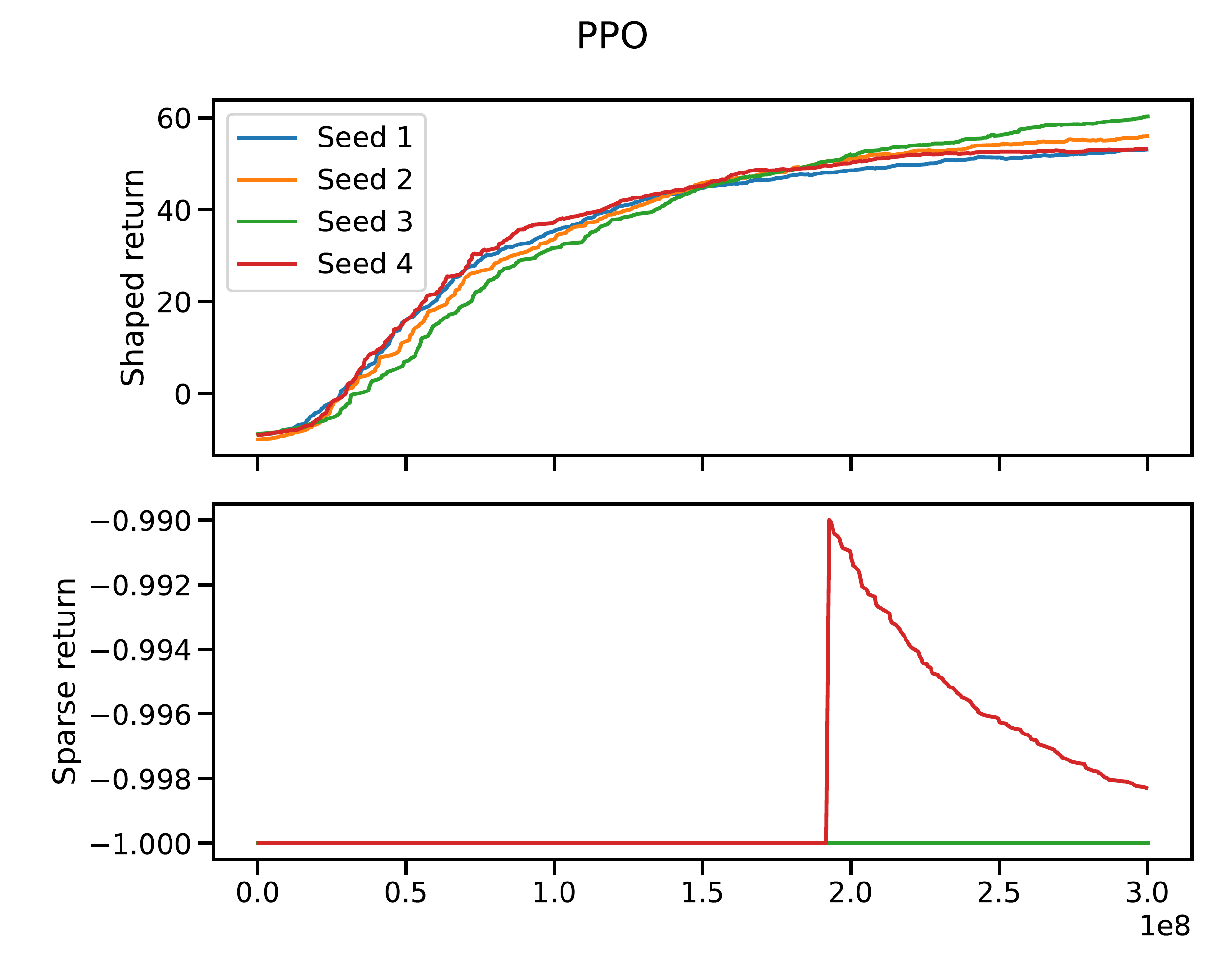}
        \caption{}
        \label{lc-ppo_coacai_no_mask}
    \end{subfigure}
    \begin{subfigure}[t]{0.5\hsize}
        \includegraphics[width=0.9\linewidth]{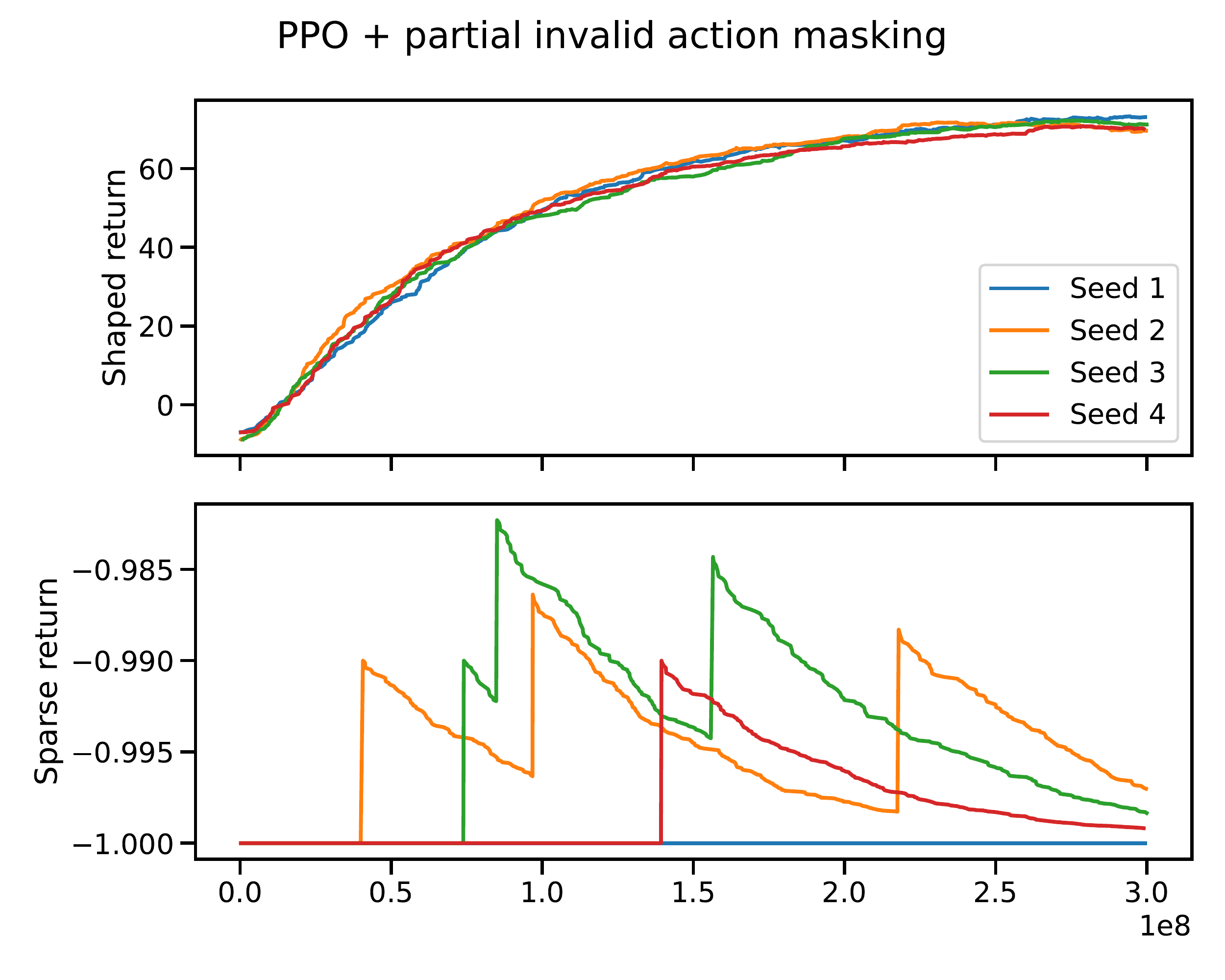}
        \caption{}
        \label{lc-ppo_coacai_partial_mask}
    \end{subfigure}
    \begin{subfigure}[t]{0.5\hsize}
        \includegraphics[width=0.9\linewidth]{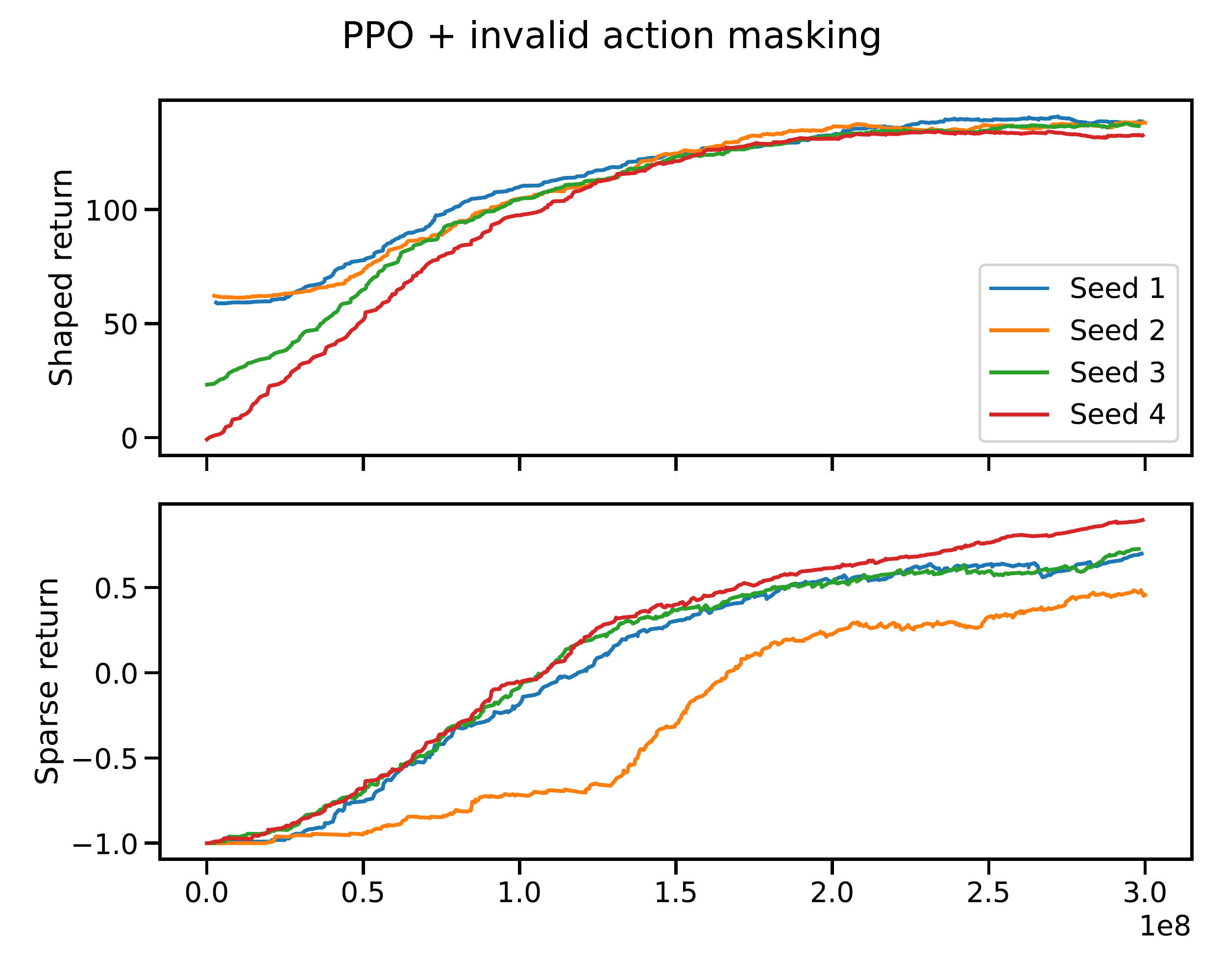}
        \caption{}
        \label{lc-ppo_coacai}
    \end{subfigure}
    \begin{subfigure}[t]{0.5\hsize}
        \includegraphics[width=0.9\linewidth]{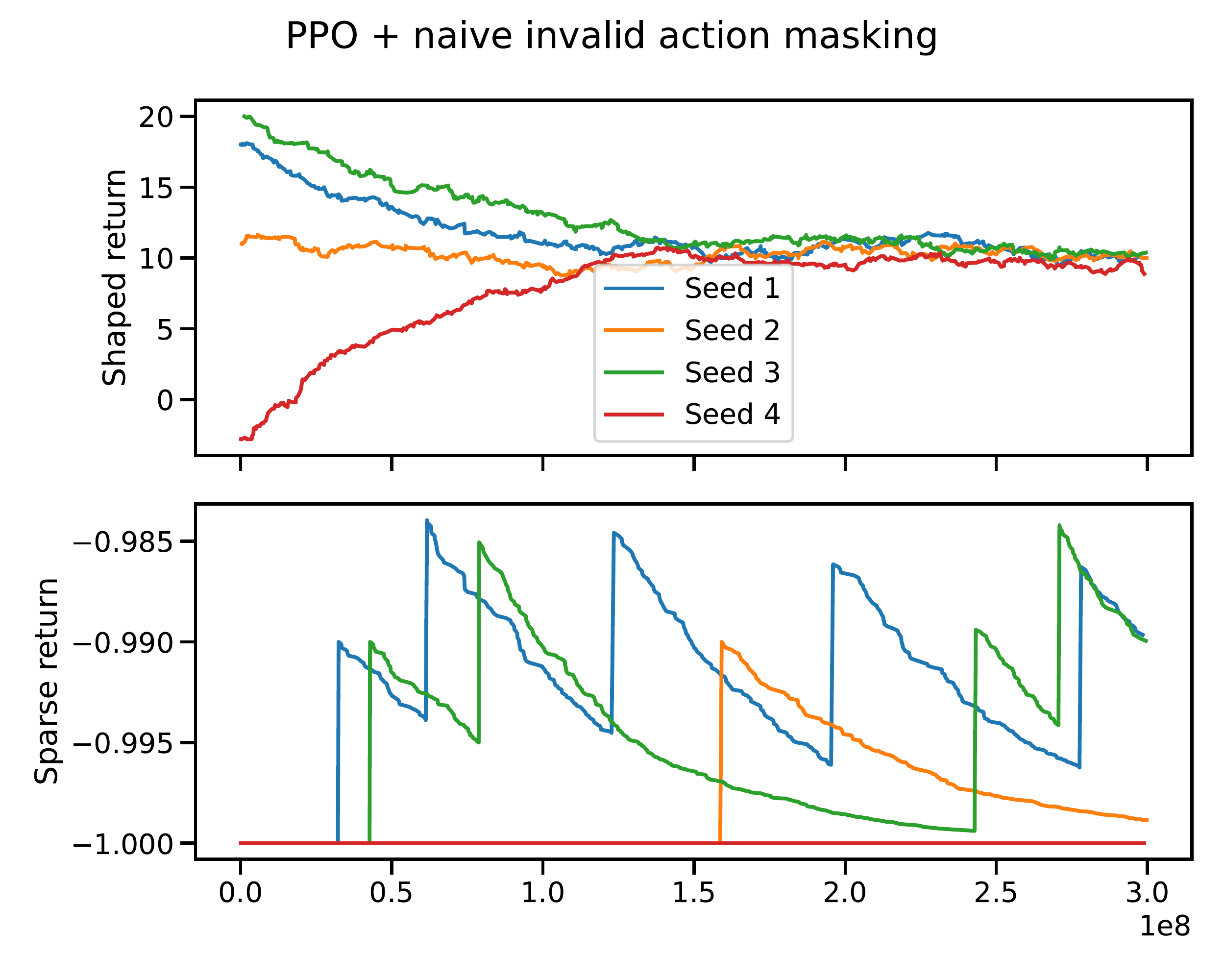}
        \caption{}
        \label{lc-ppo_coacai_naive}
    \end{subfigure}   
    \begin{subfigure}[t]{0.5\hsize}
        \includegraphics[width=0.9\linewidth]{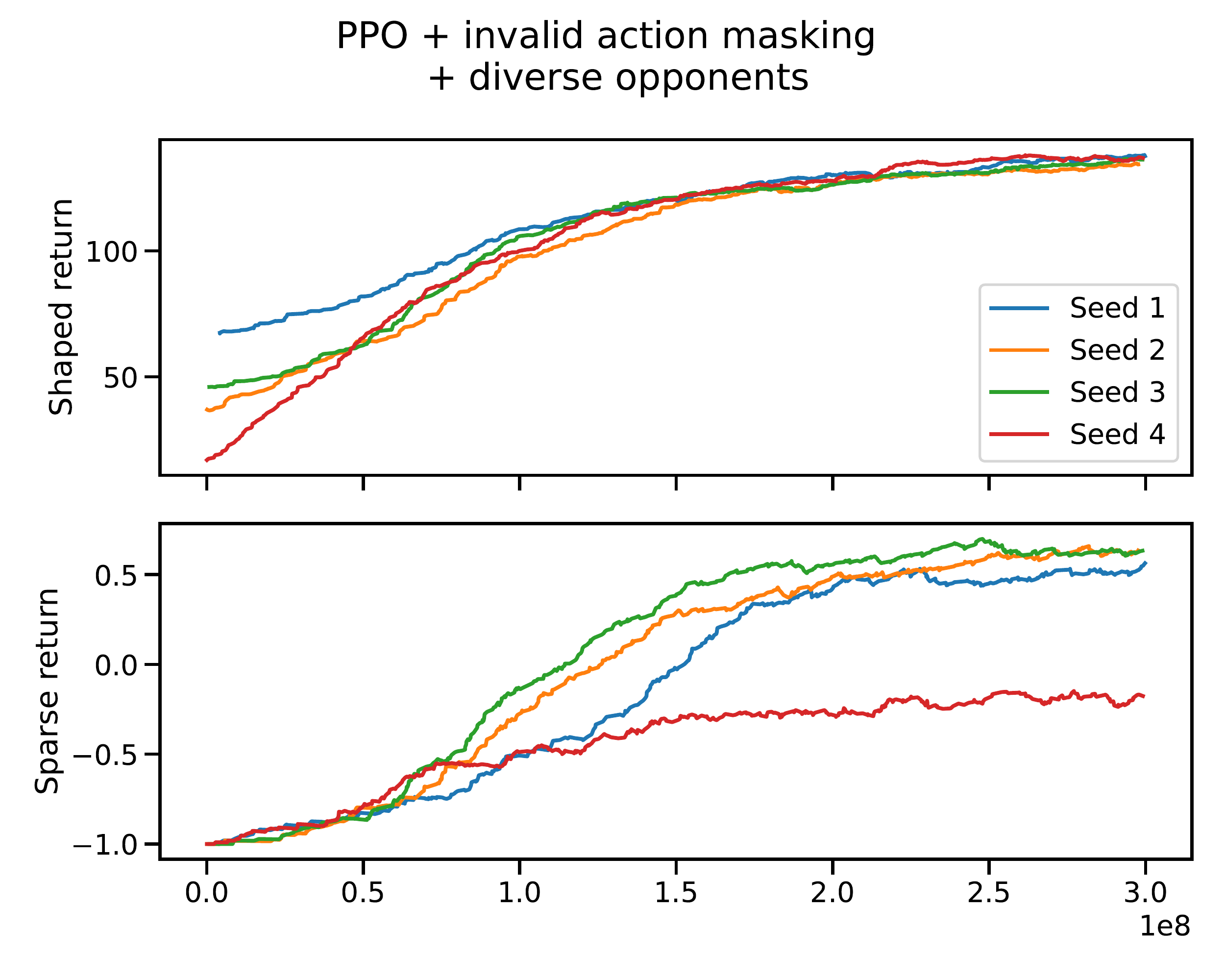}
        \caption{}
        \label{lc-ppo_diverse}
    \end{subfigure}
    \begin{subfigure}[t]{0.5\hsize}
        \includegraphics[width=0.9\linewidth]{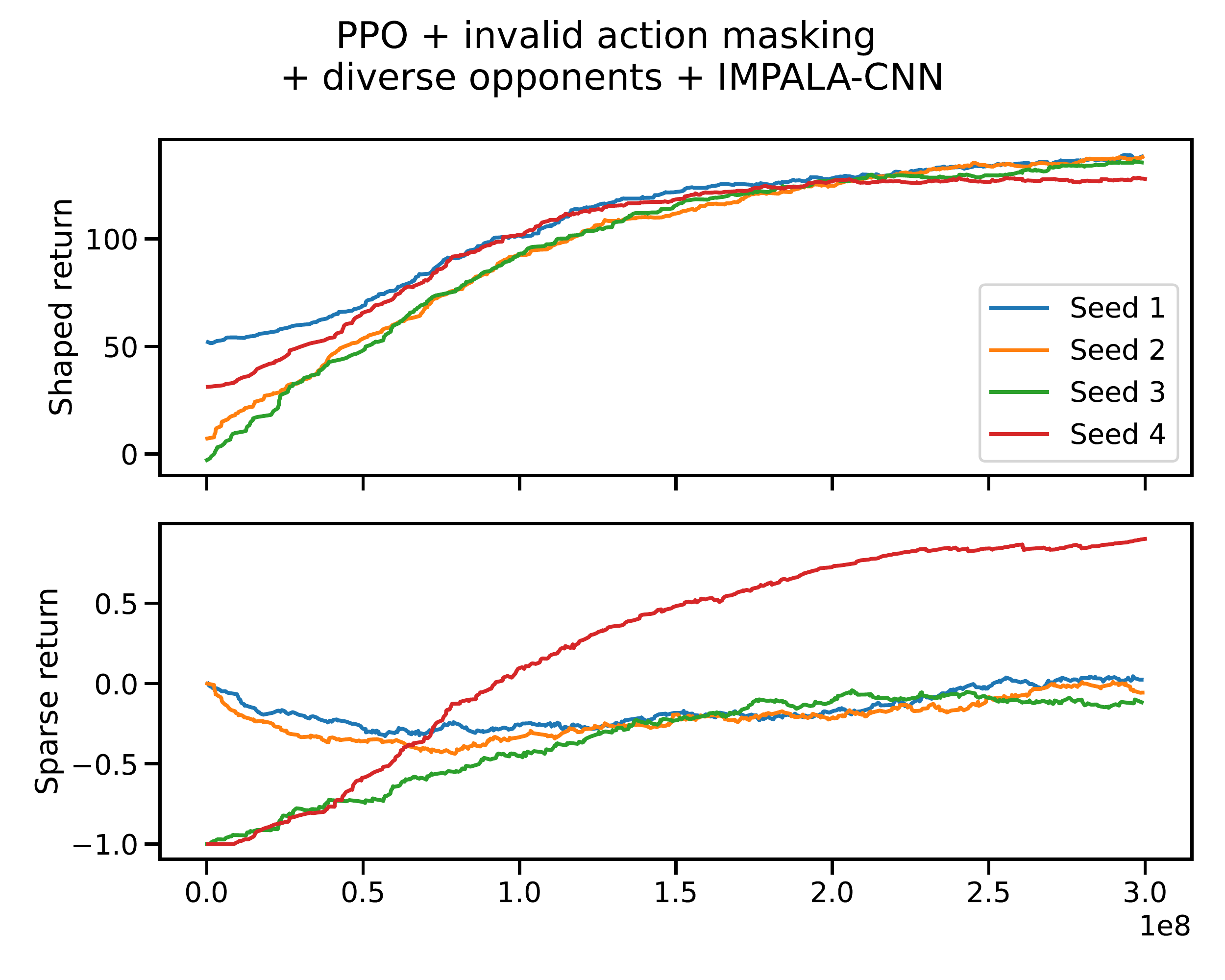}
        \caption{}
        \label{lc-ppo_diverse_impala}
    \end{subfigure}
    \caption{UAS learning curves.}
    \label{fig:UAS-learning-curves}
\end{figure}

\begin{figure}[h]
    \begin{subfigure}[t]{0.5\hsize}
        \includegraphics[width=0.9\linewidth]{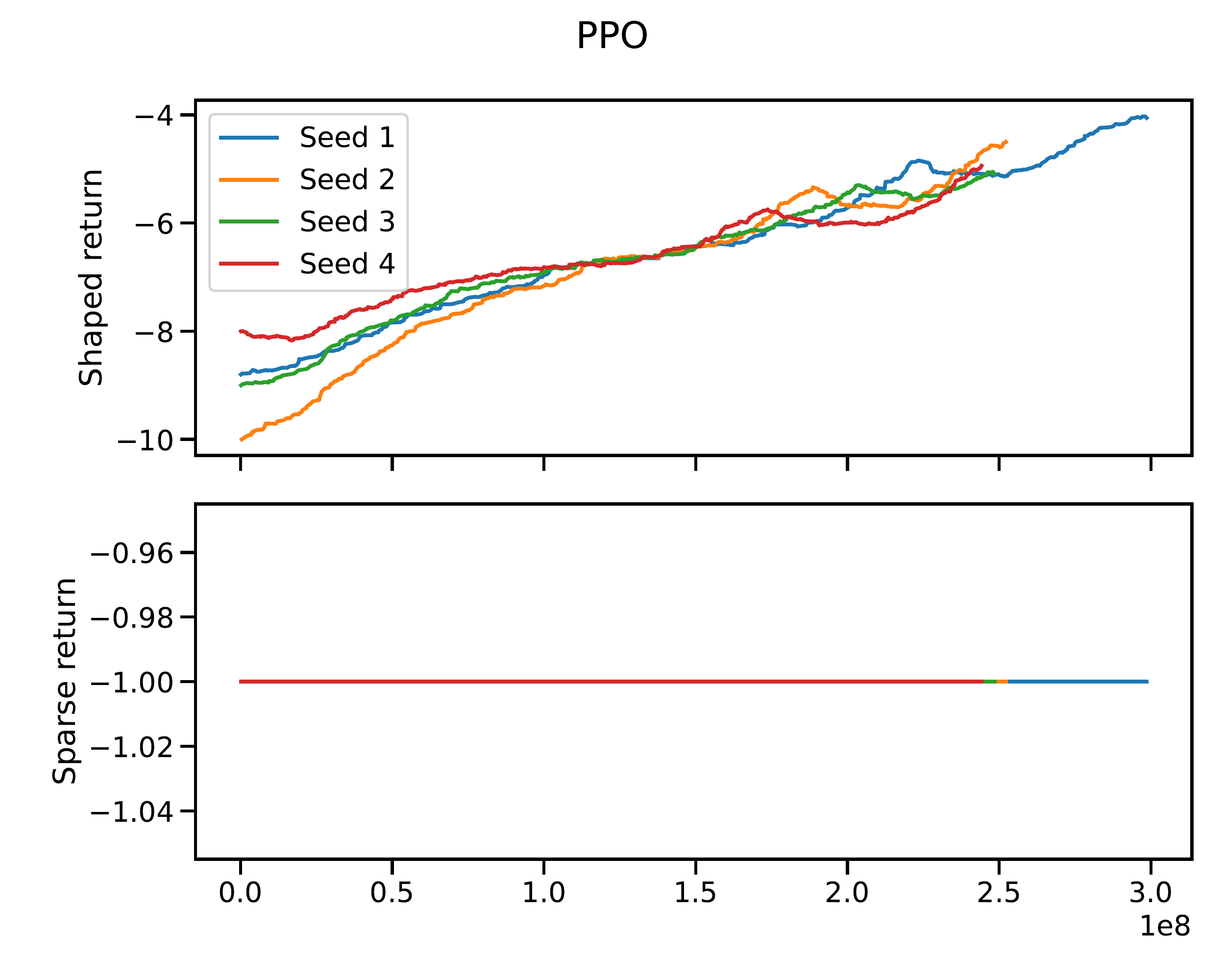}
        \caption{}
        \label{lc-ppo_coacai_naive}
    \end{subfigure}   
    \begin{subfigure}[t]{0.5\hsize}
        \includegraphics[width=0.9\linewidth]{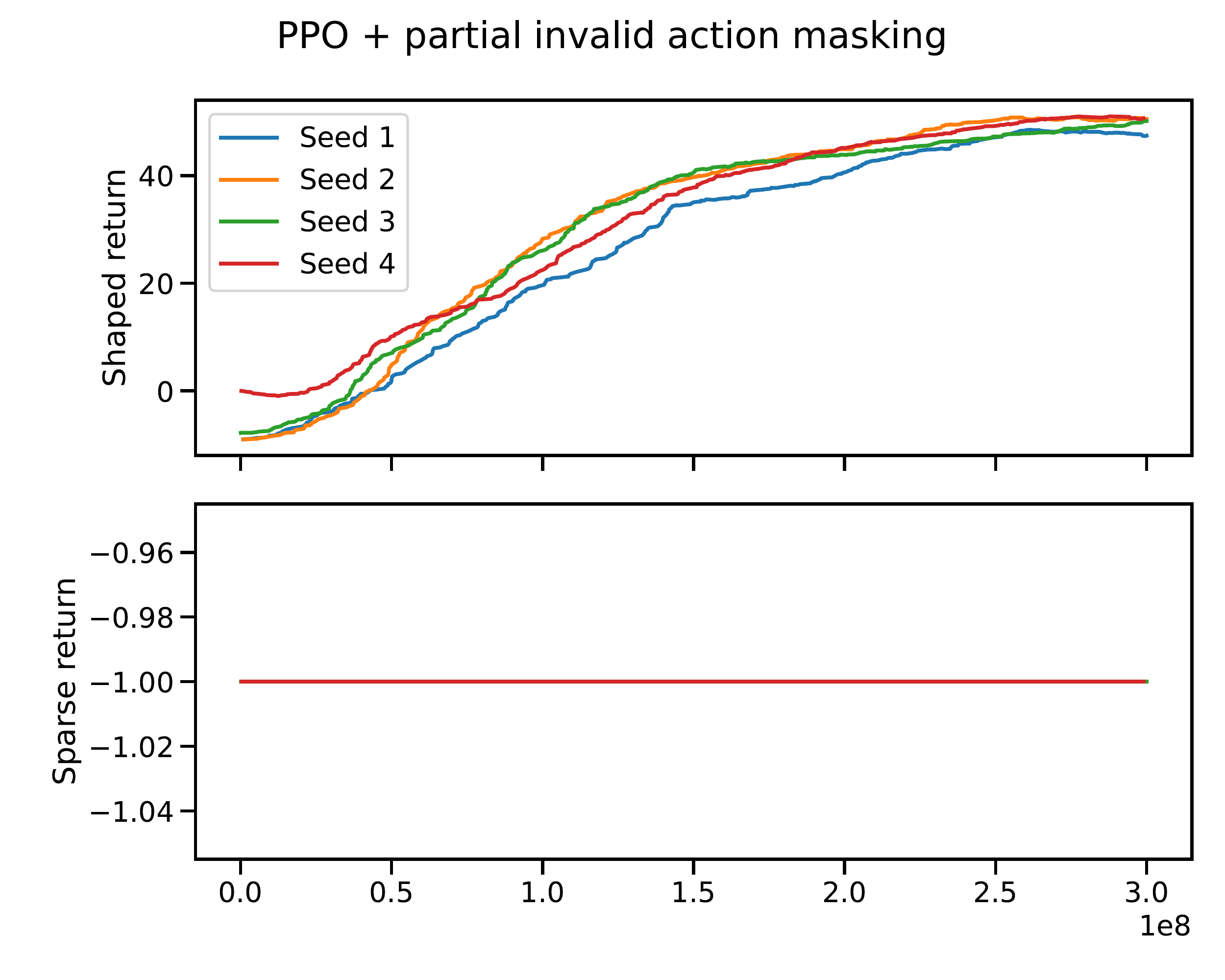}
        \caption{}
        \label{lc-ppo_coacai_no_mask}
    \end{subfigure}
    \begin{subfigure}[t]{0.5\hsize}
        \includegraphics[width=0.9\linewidth]{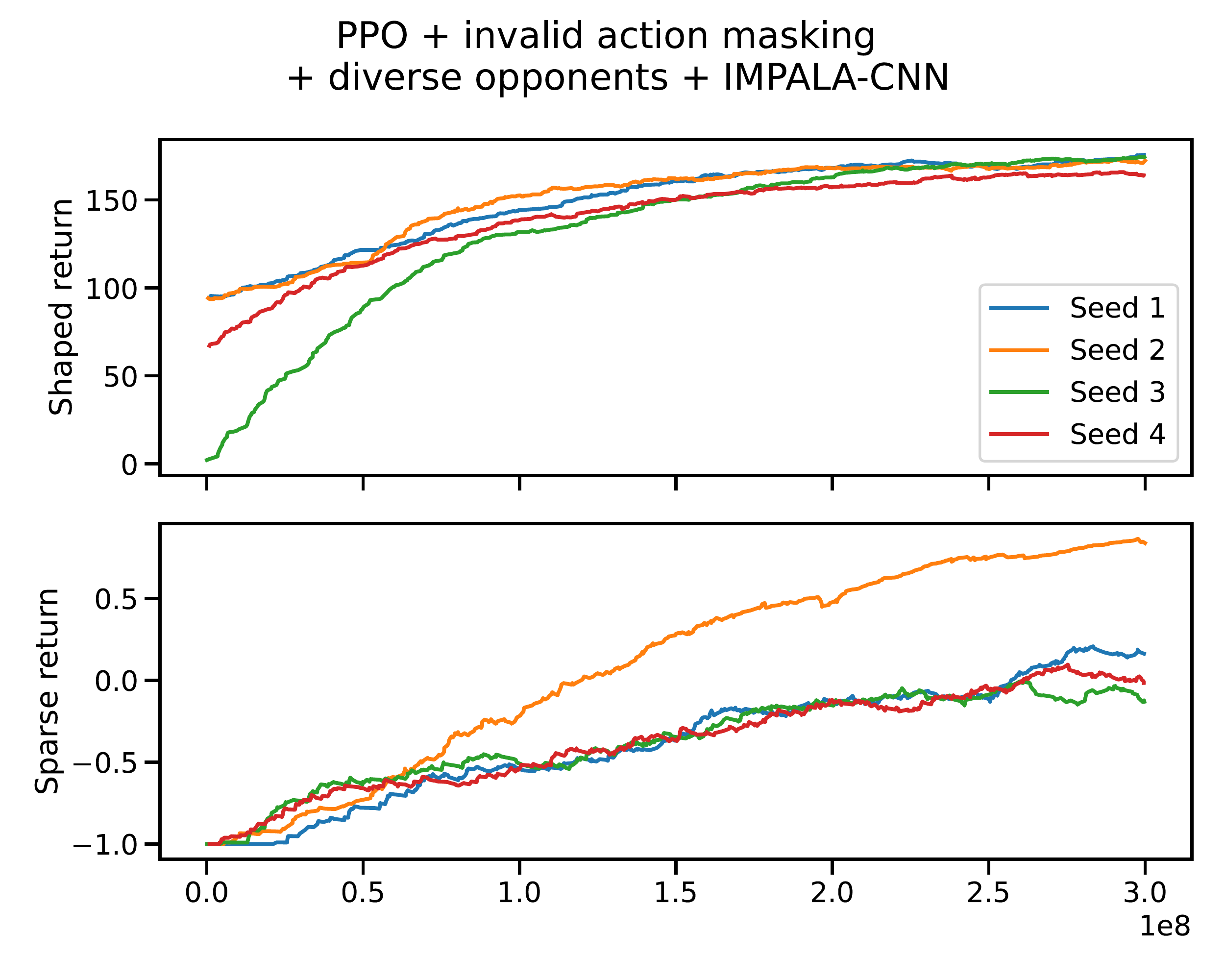}
        \caption{}
        \label{lc-ppo_coacai_partial_mask}
    \end{subfigure}
    \begin{subfigure}[t]{0.5\hsize}
        \includegraphics[width=0.9\linewidth]{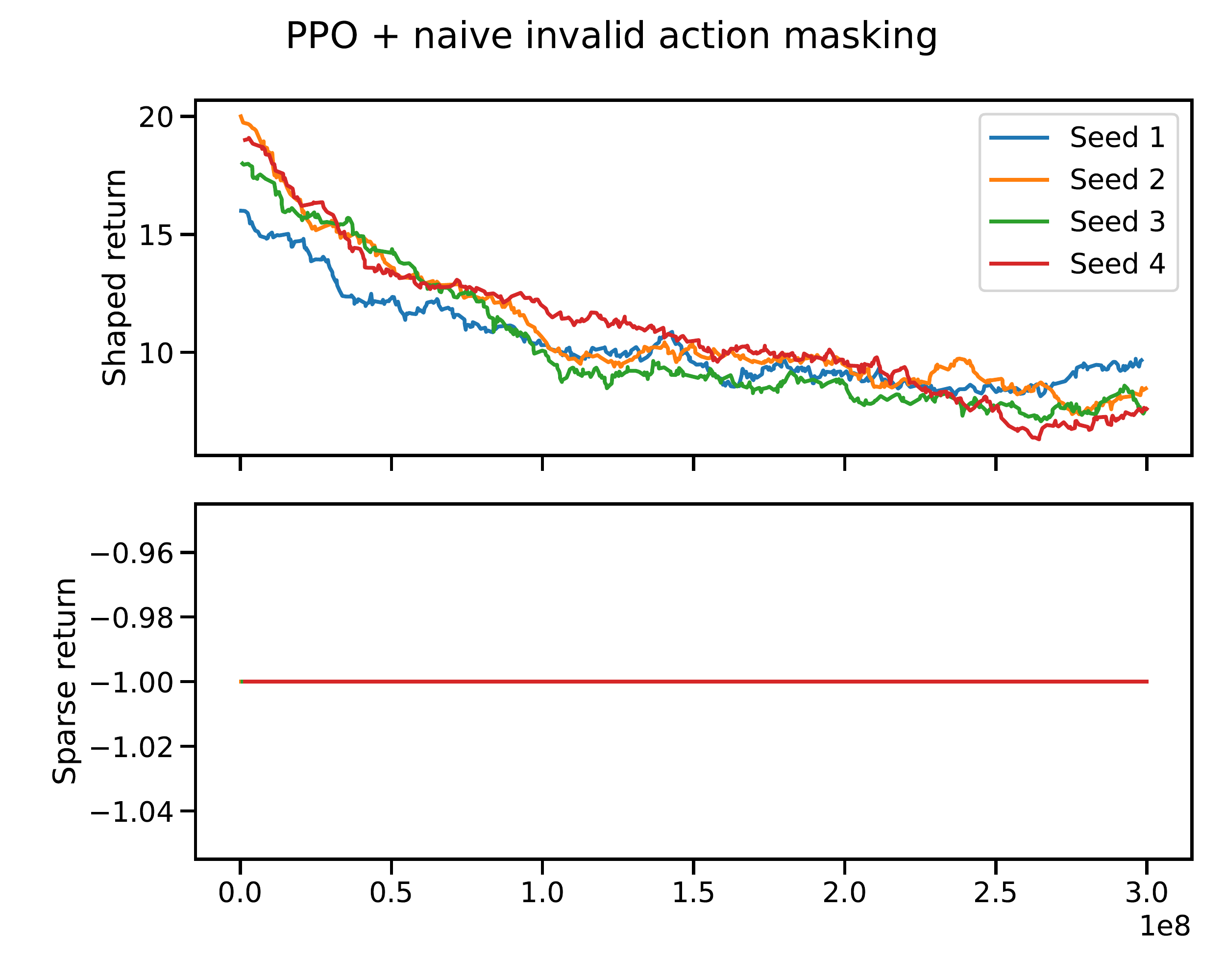}
        \caption{}
        \label{lc-ppo_coacai_naive}
    \end{subfigure}   
    \begin{subfigure}[t]{0.5\hsize}
        \includegraphics[width=0.9\linewidth]{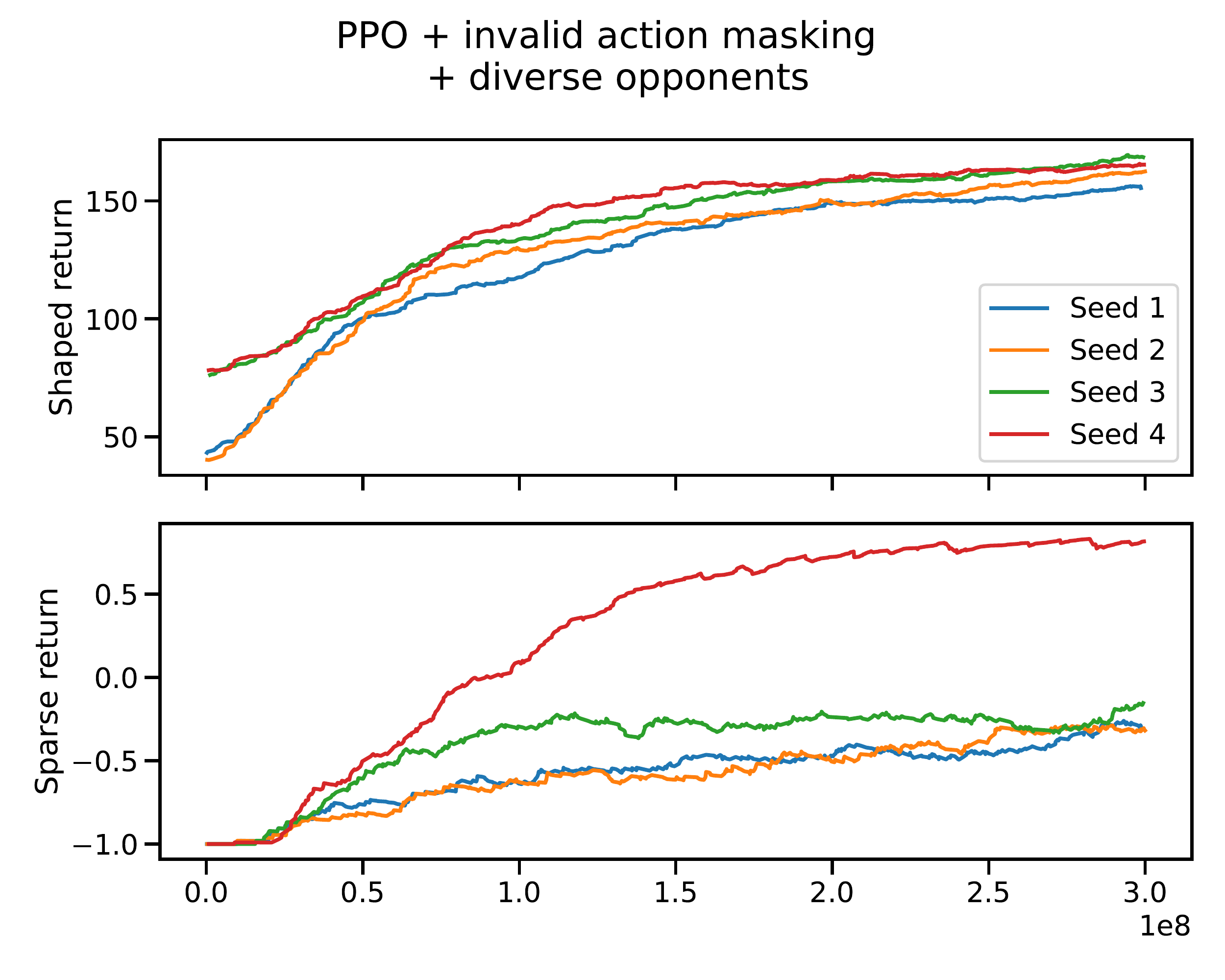}
        \caption{}
        \label{lc-ppo_coacai}
    \end{subfigure}
    \begin{subfigure}[t]{0.5\hsize}
        \includegraphics[width=0.9\linewidth]{charts/shaped_vs_sparse/ppo_gridnet_diverse_impala.pdf}
        \caption{}
        \label{lc-ppo_coacai}
    \end{subfigure}
    \caption{Gridnet learning curves.}
    \label{fig:Gridnet-learning-curves}
\end{figure}

All the learning curves related for UAS and Gridnet can be found at Figure~\ref{fig:UAS-learning-curves} and Figure~\ref{fig:Gridnet-learning-curves}, respectively. Similarly, the match results can be found at Figure~\ref{fig:UAS-match-results} and Figure~\ref{fig:Gridnet-match-results}.


\begin{figure}[h]
    \begin{subfigure}[t]{0.5\hsize}
        \includegraphics[width=0.9\linewidth]{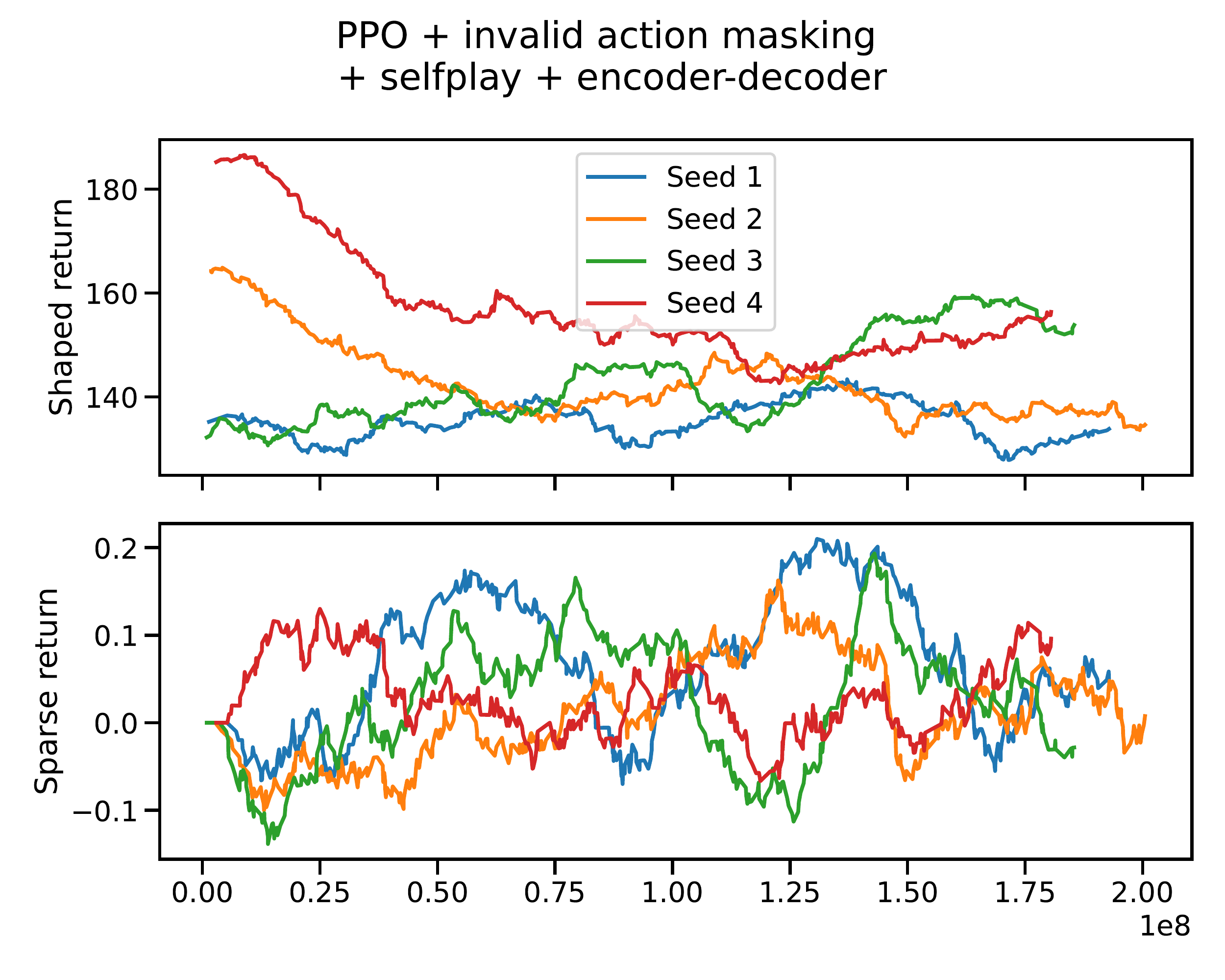}
        \caption{}
        \label{lc-ppo_coacai_naive}
    \end{subfigure}   
    \begin{subfigure}[t]{0.5\hsize}
        \includegraphics[width=0.9\linewidth]{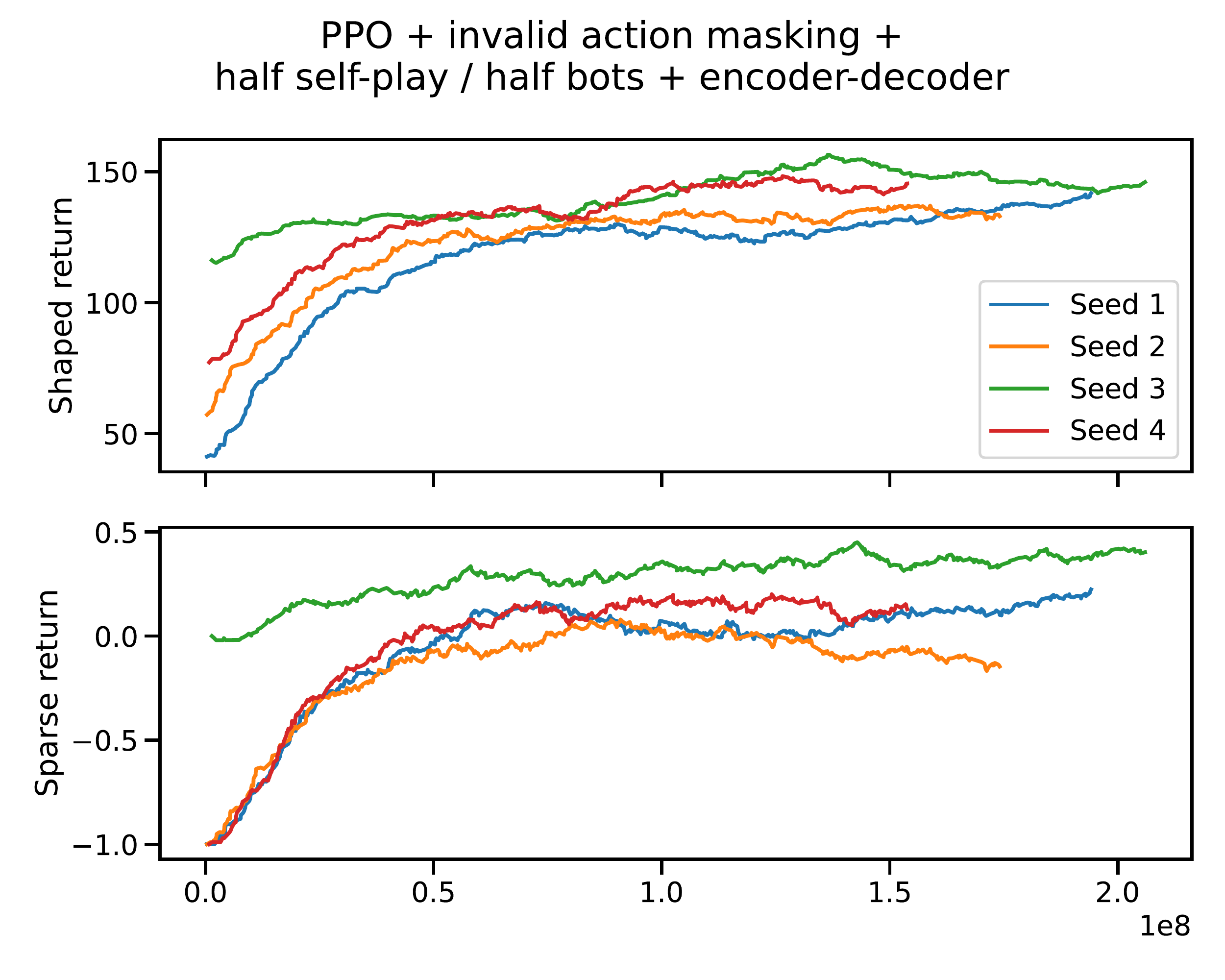}
        \caption{}
        \label{lc-ppo_coacai_no_mask}
    \end{subfigure}
    \caption{Gridnet selfplay learning curves.}
    \label{fig:Gridnet-selfplay-learning-curves}
\end{figure}





\begin{figure}[h]
    \begin{subfigure}[t]{0.5\hsize}
        \includegraphics[width=0.9\linewidth]{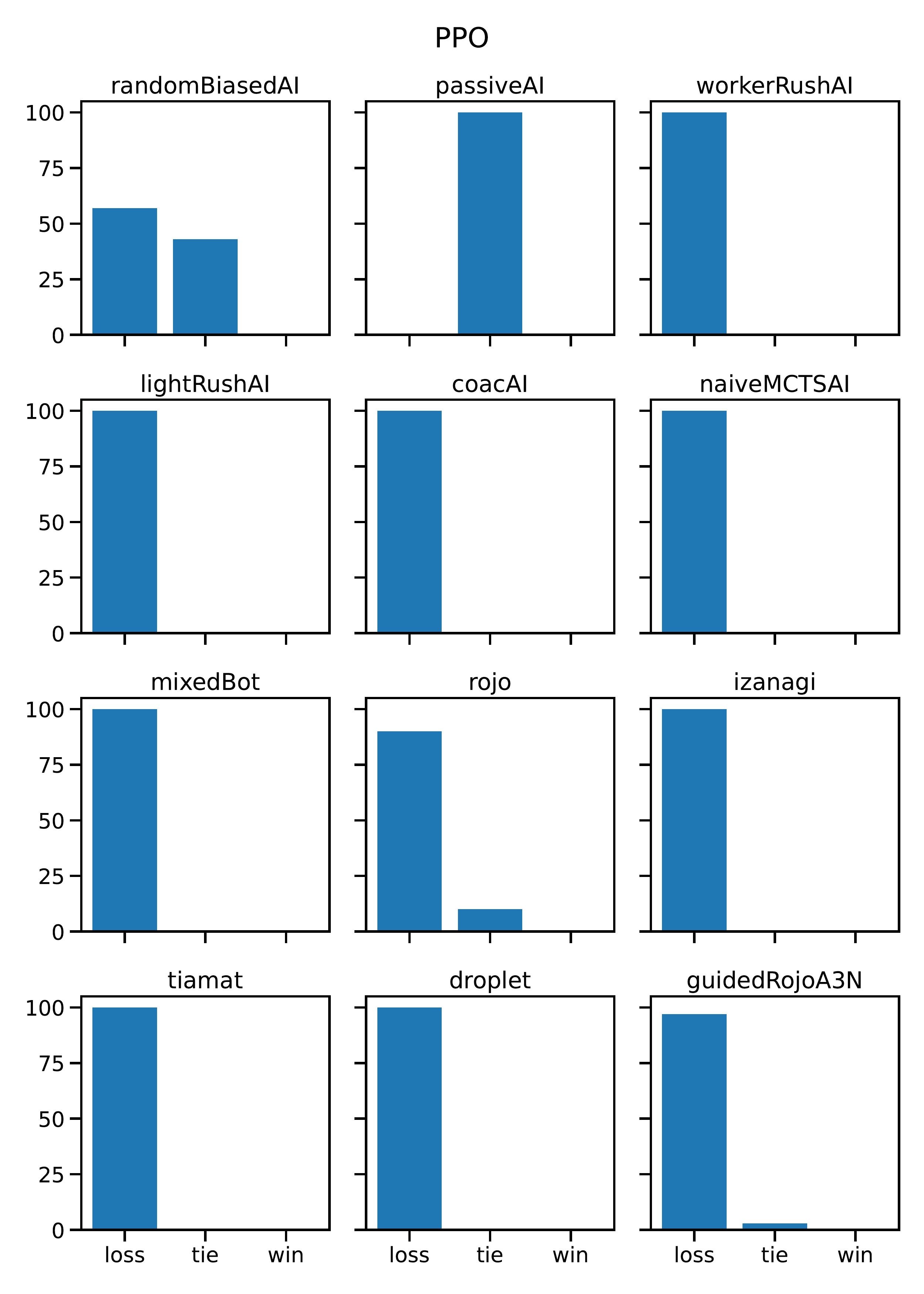}
        \caption{}
    \end{subfigure}
    \begin{subfigure}[t]{0.5\hsize}
        \includegraphics[width=0.9\linewidth]{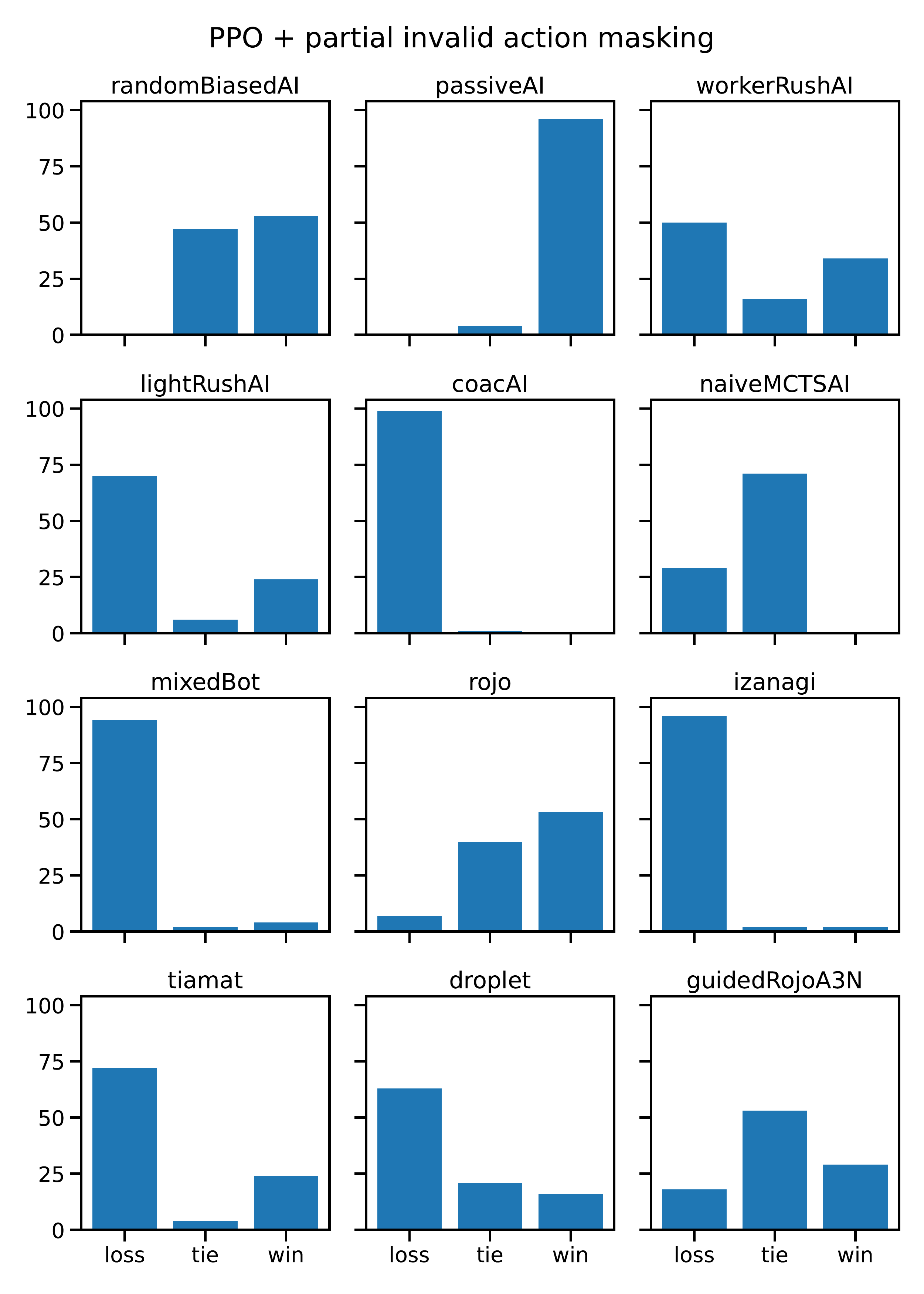}
        \caption{}
    \end{subfigure}
    \begin{subfigure}[t]{0.5\hsize}
        \includegraphics[width=0.9\linewidth]{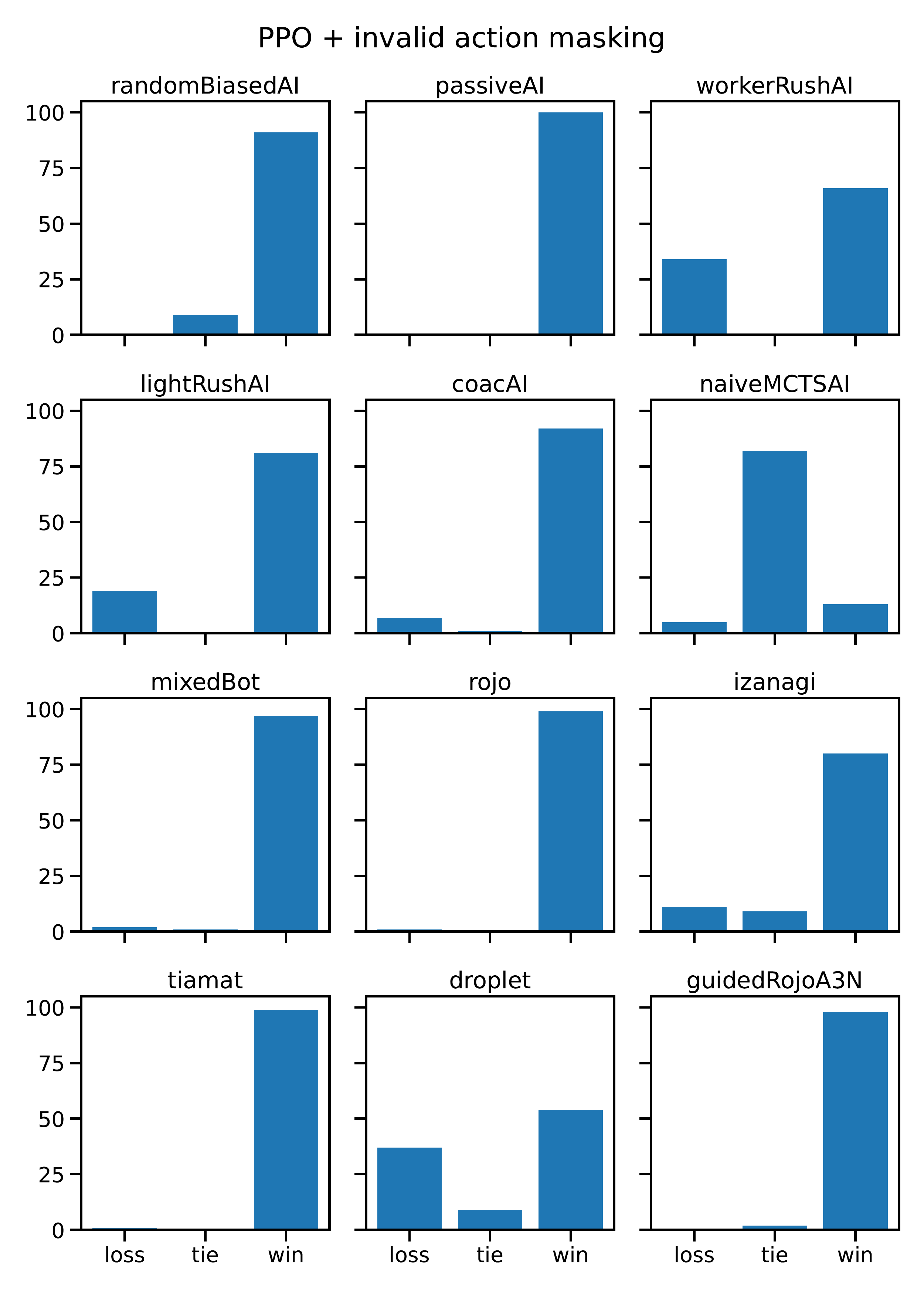}
        \caption{}
    \end{subfigure}
    \begin{subfigure}[t]{0.5\hsize}
        \includegraphics[width=0.9\linewidth]{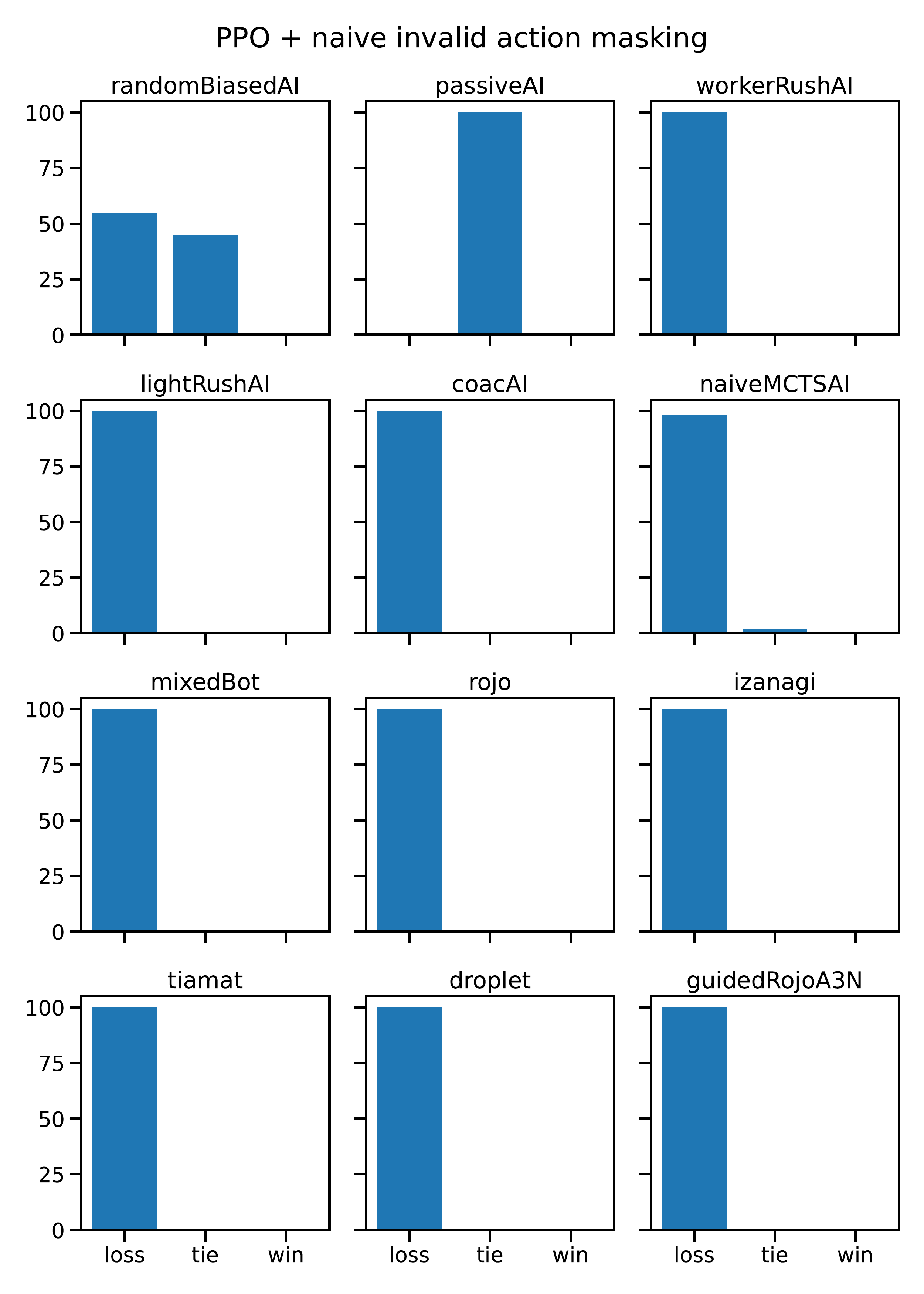}
        \caption{}
    \end{subfigure}  
        \caption{UAS match results.}
        \end{figure}
    \clearpage   
        \begin{figure}[tb]\ContinuedFloat
    \begin{subfigure}[t]{0.5\hsize}
        \includegraphics[width=0.9\linewidth]{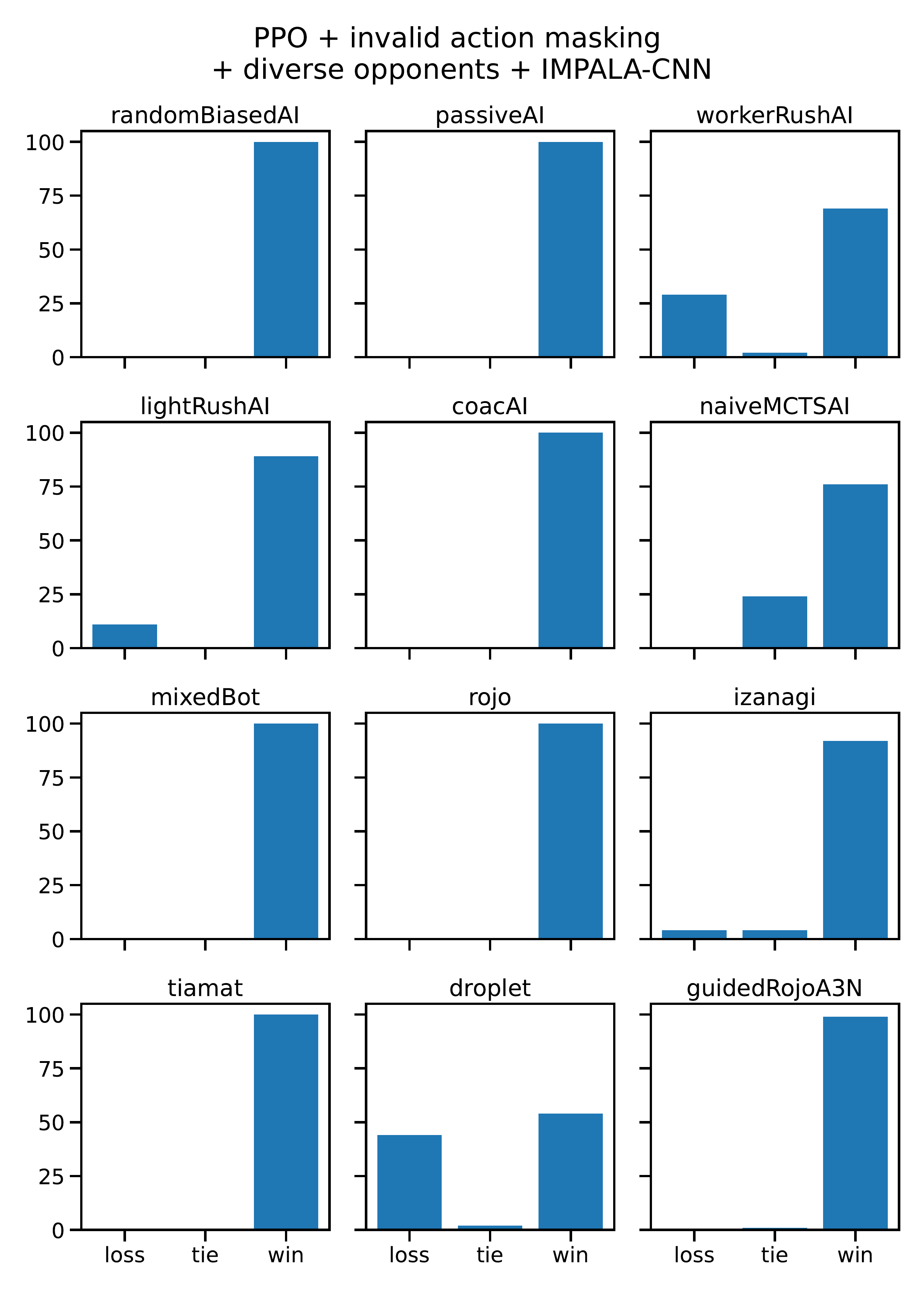}
        \caption{}
    \end{subfigure}
    \begin{subfigure}[t]{0.5\hsize}
        \includegraphics[width=0.9\linewidth]{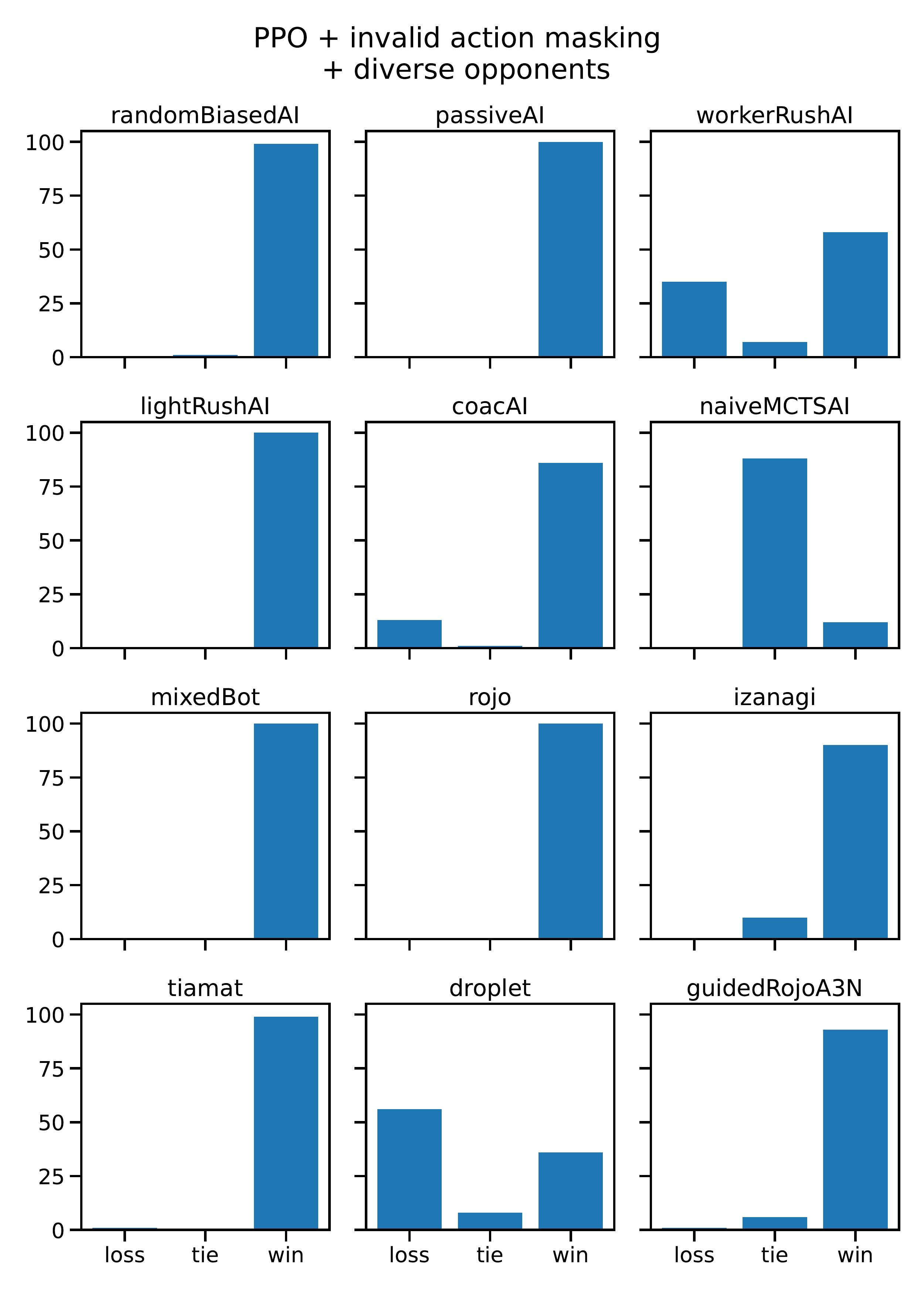}
        \caption{}
    \end{subfigure}
    \caption{UAS match results.}
    \label{fig:UAS-match-results}
\end{figure}

\begin{figure}[h]
    \begin{subfigure}[t]{0.5\hsize}
        \includegraphics[width=0.9\linewidth]{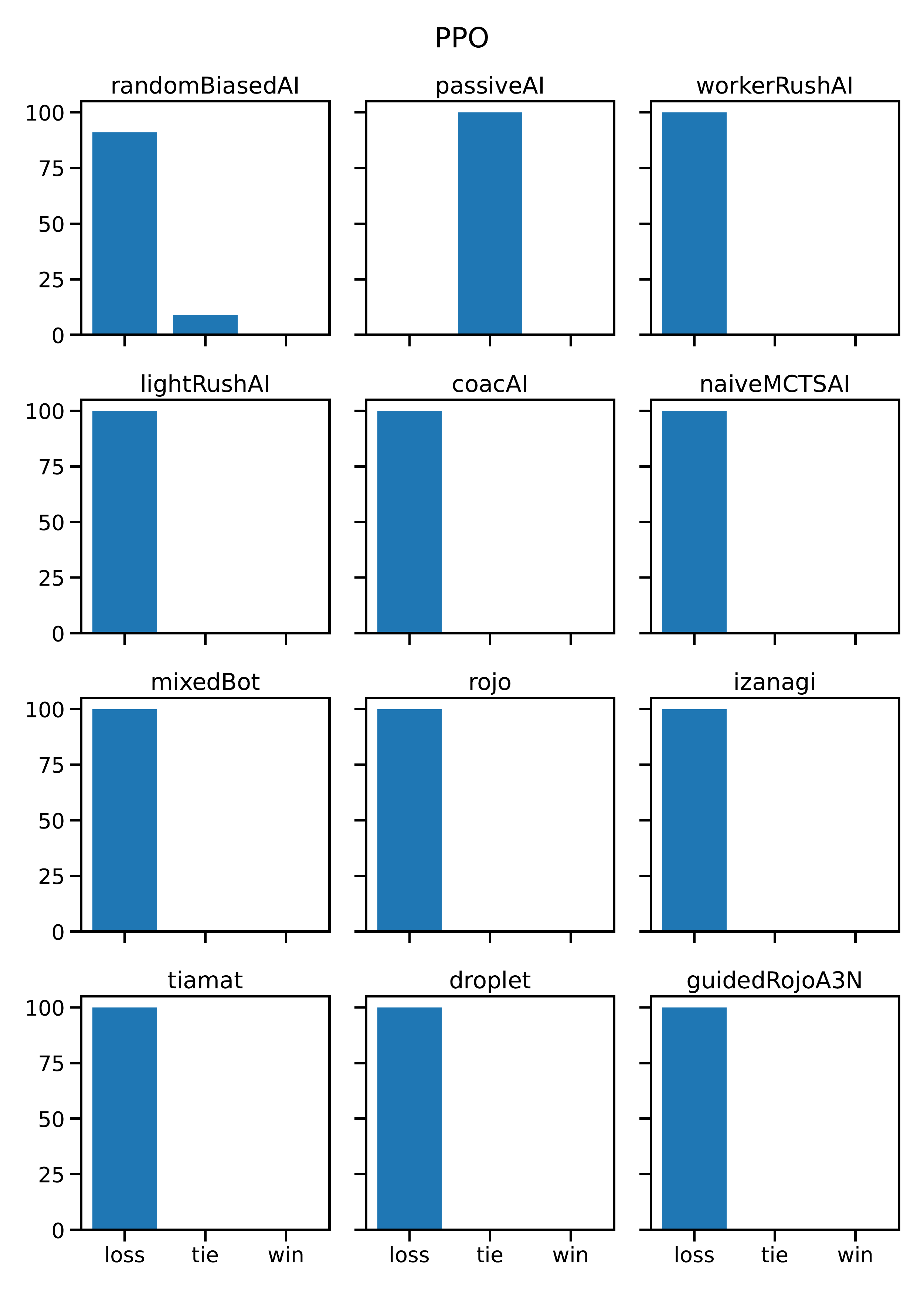}
        \caption{}
    \end{subfigure}
    \begin{subfigure}[t]{0.5\hsize}
        \includegraphics[width=0.9\linewidth]{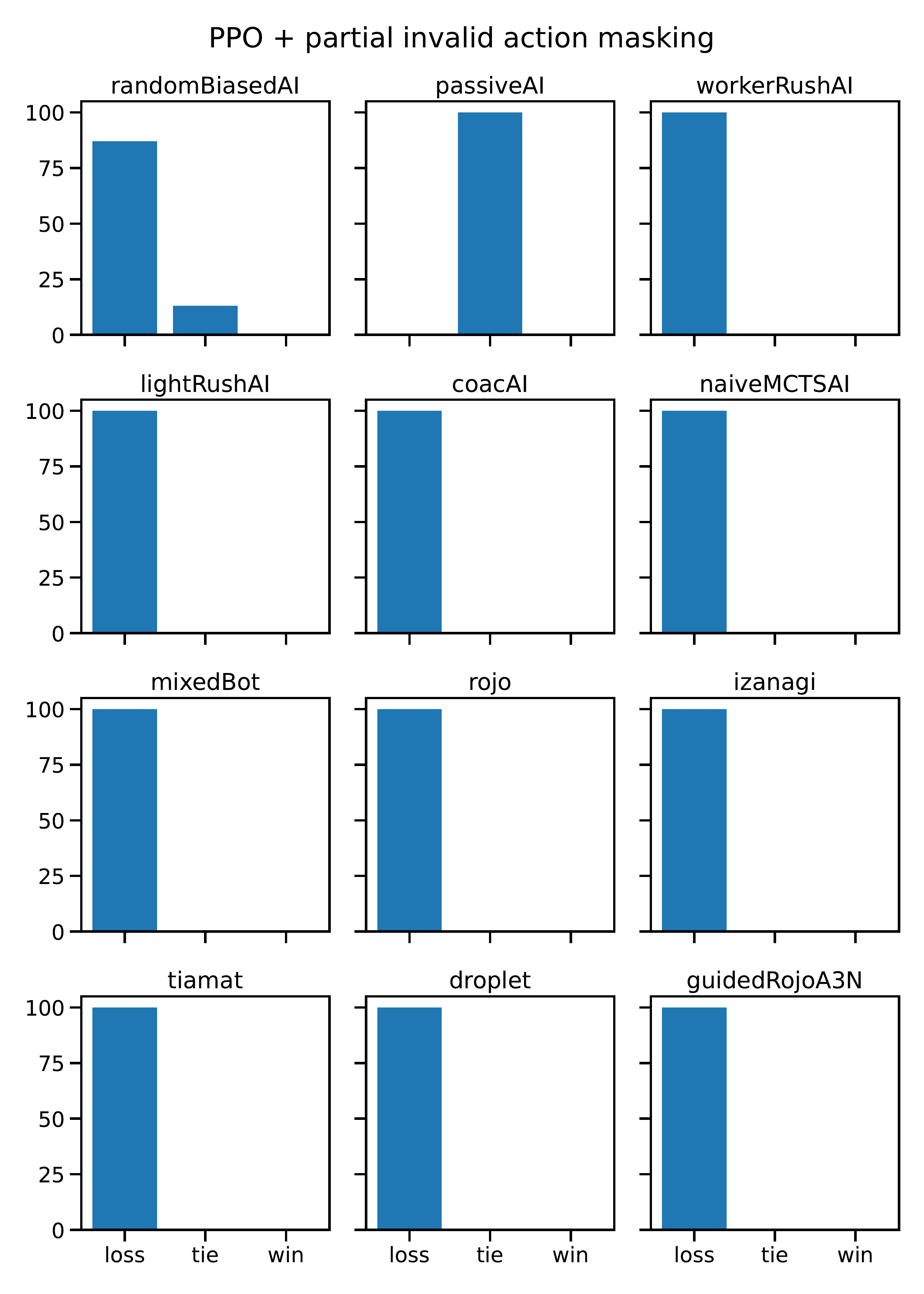}
        \caption{}
    \end{subfigure}
    \begin{subfigure}[t]{0.5\hsize}
        \includegraphics[width=0.9\linewidth]{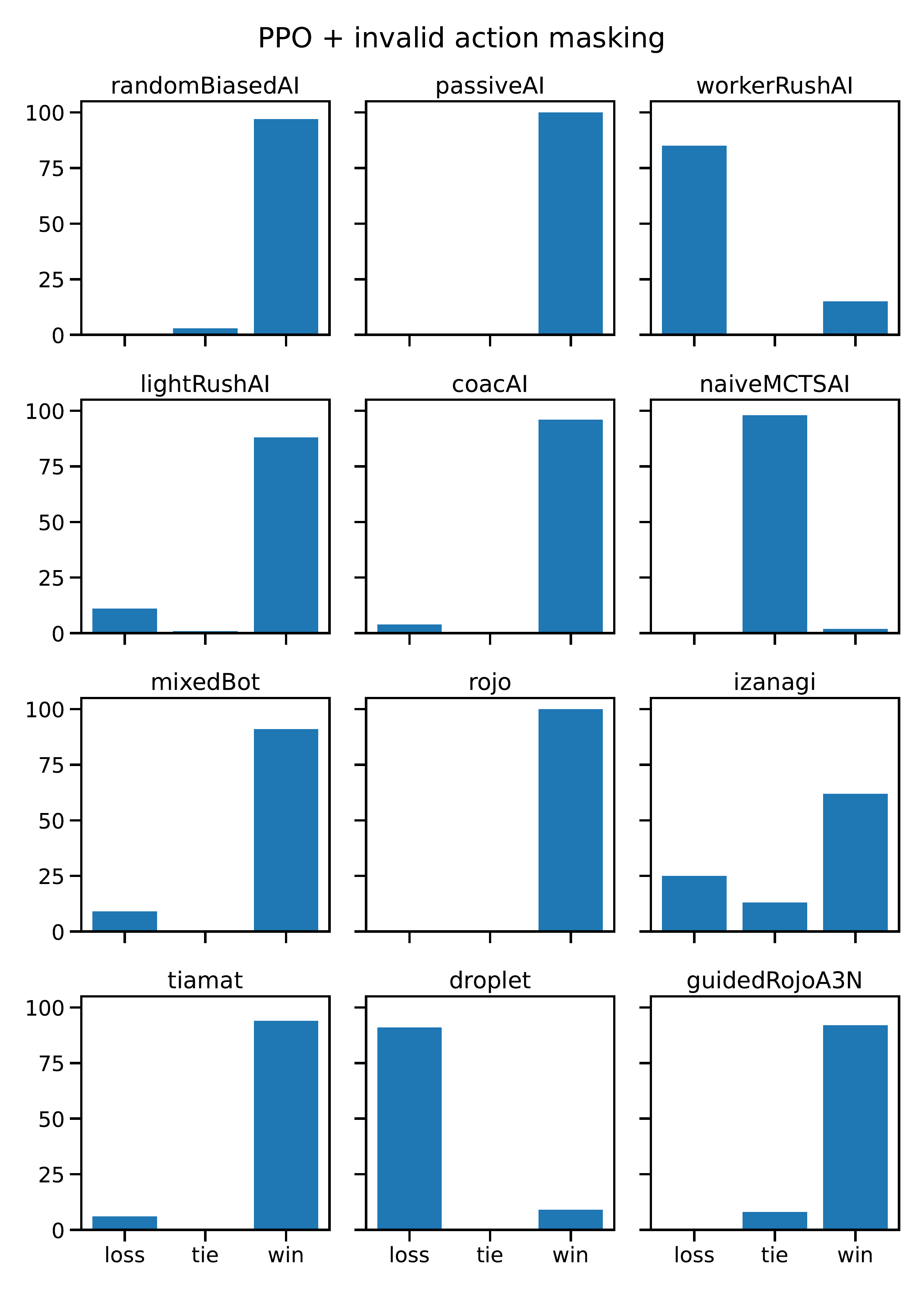}
        \caption{}
    \end{subfigure}
    \begin{subfigure}[t]{0.5\hsize}
        \includegraphics[width=0.9\linewidth]{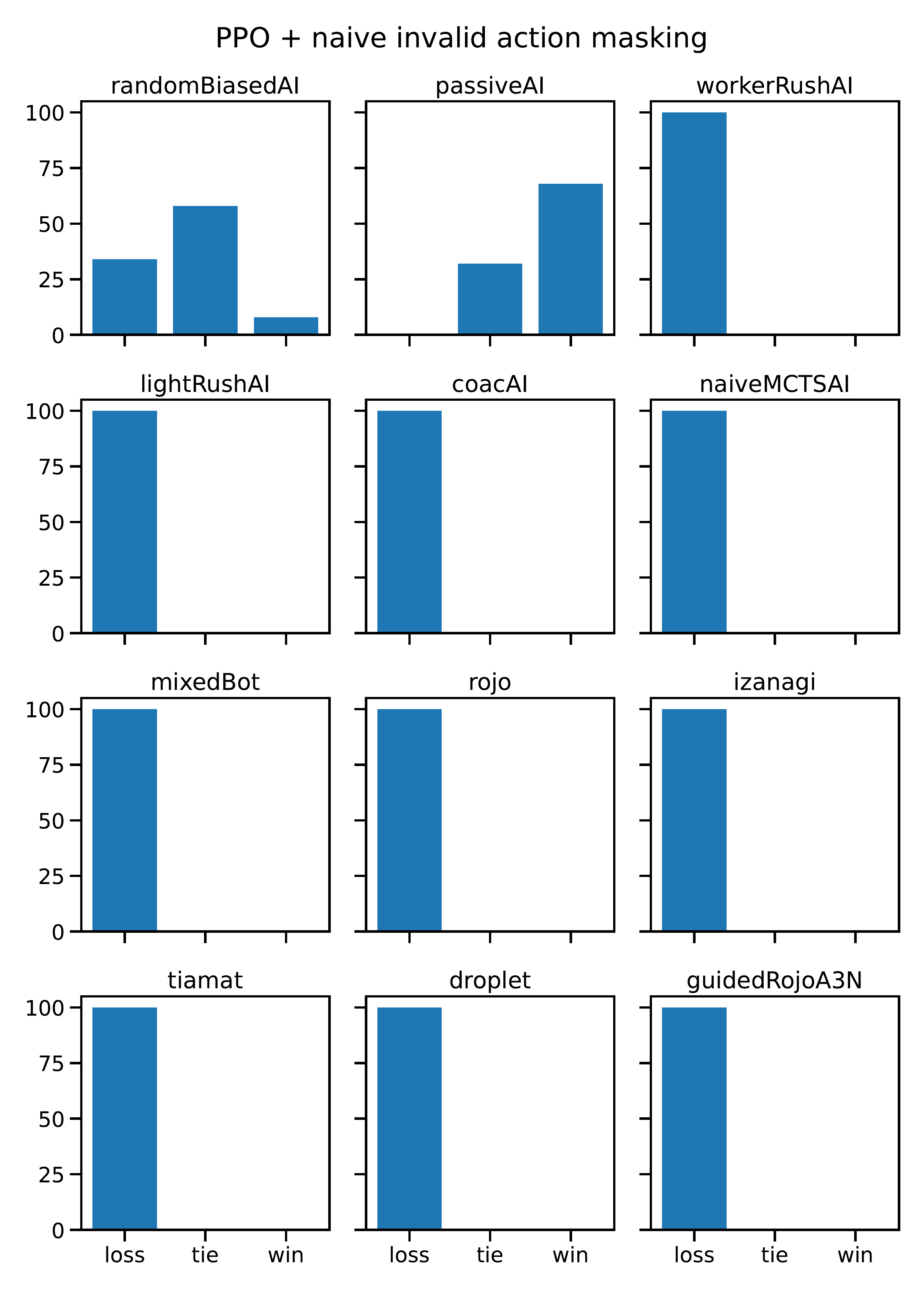}
        \caption{}
    \end{subfigure}   
        \caption{Gridnet match results.}
        \end{figure}
    \clearpage   
        \begin{figure}[tb]\ContinuedFloat
    \begin{subfigure}[t]{0.5\hsize}
        \includegraphics[width=0.9\linewidth]{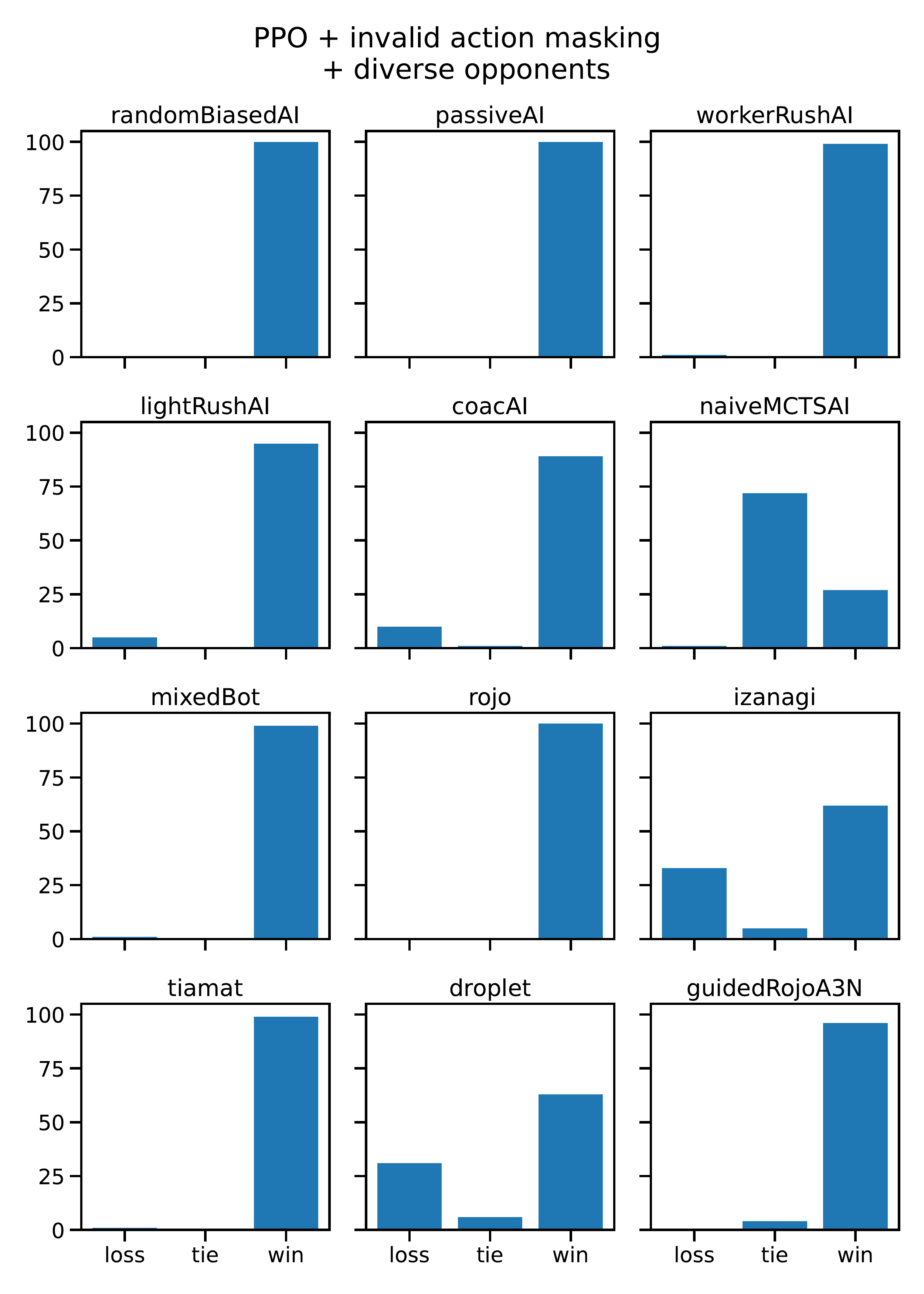}
        \caption{}
    \end{subfigure}
    \begin{subfigure}[t]{0.5\hsize}
        \includegraphics[width=0.9\linewidth]{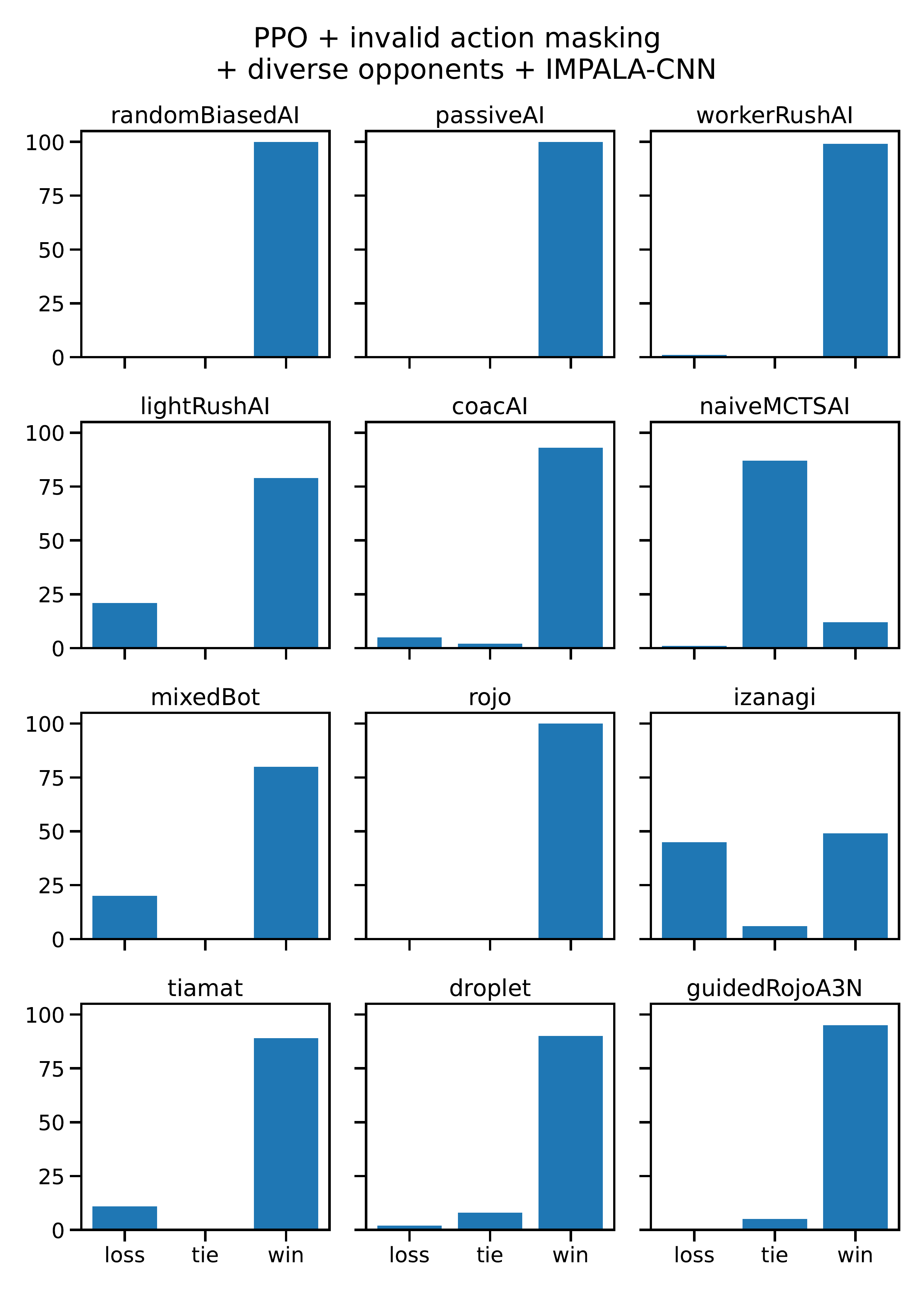}
        \caption{}
    \end{subfigure}
    \begin{subfigure}[t]{0.5\hsize}
        \includegraphics[width=0.9\linewidth]{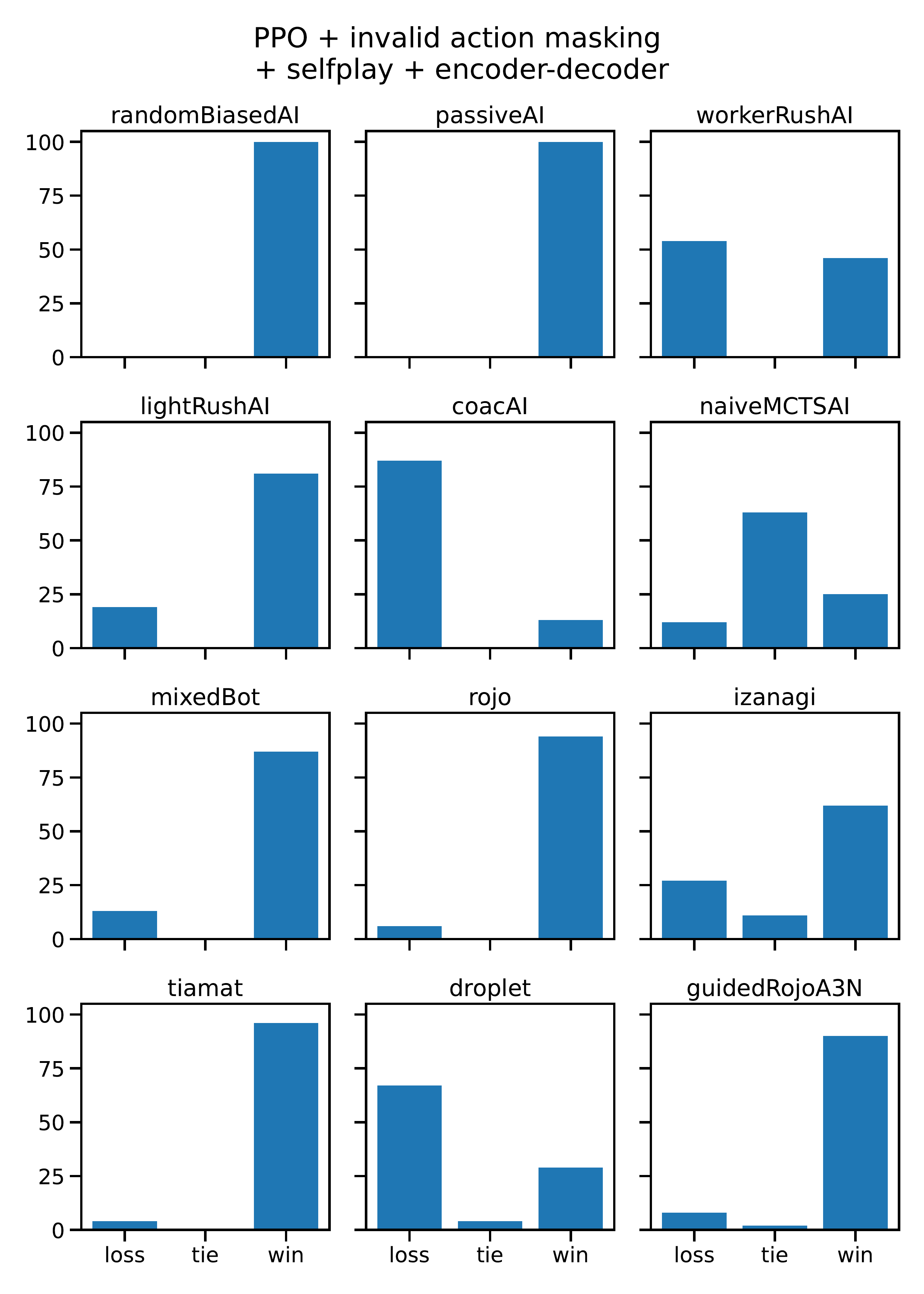}
        \caption{}
    \end{subfigure}
    \begin{subfigure}[t]{0.5\hsize}
        \includegraphics[width=0.9\linewidth]{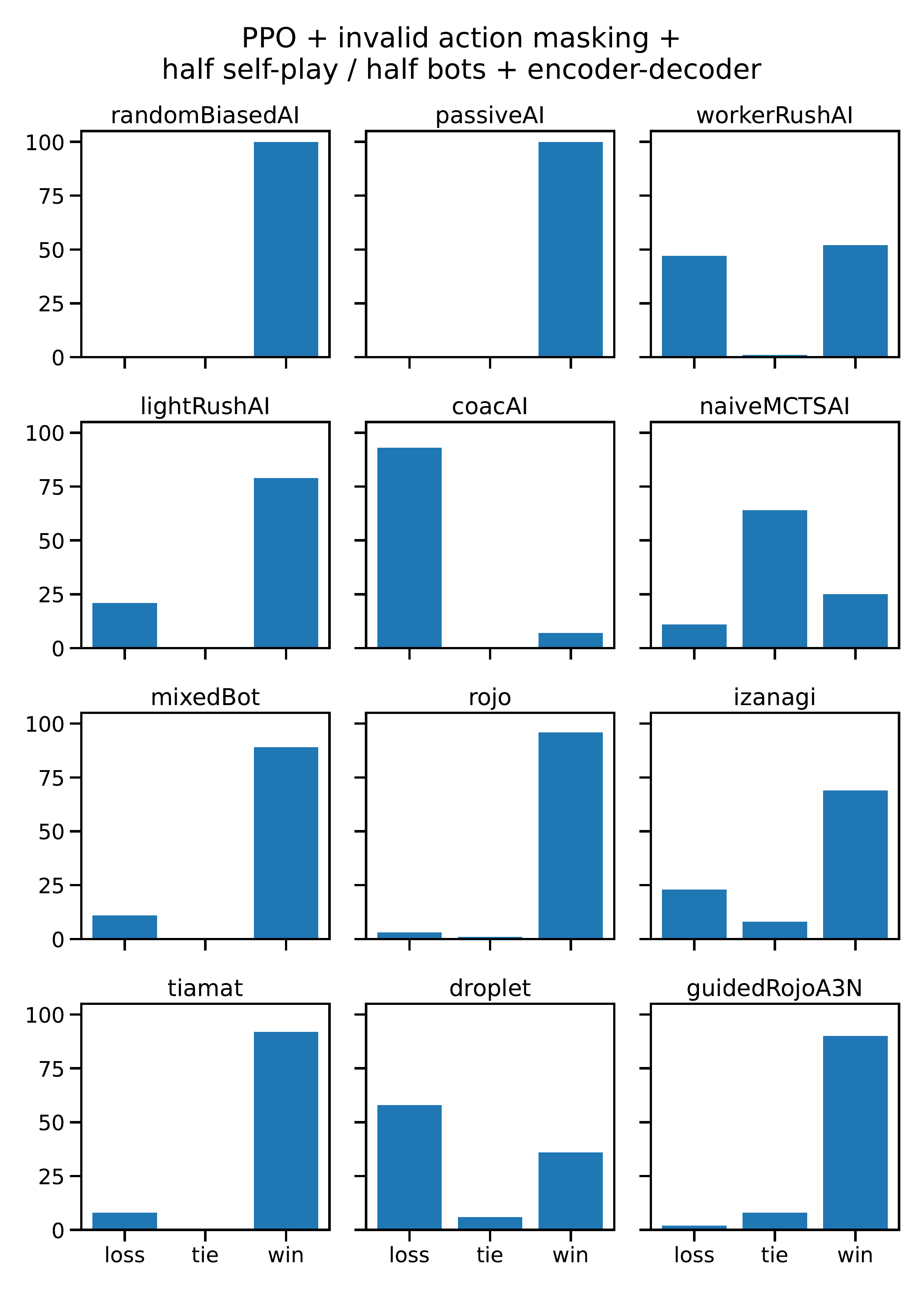}
        \caption{}
    \end{subfigure}
    \caption{Gridnet match results.}
    \label{fig:Gridnet-match-results}
\end{figure}

\end{document}